\documentclass[journal]{IEEEtran}
 \usepackage[pdftex]{graphicx}
\usepackage{subfigure}
\usepackage{amsmath,amssymb,amsfonts}
\usepackage{color}
\usepackage{blindtext}
\usepackage{algorithm}
\usepackage{algorithmic}
\usepackage{cite}
\usepackage{amsthm}
\usepackage[normalem]{ulem}
\usepackage{url}
\usepackage{comment}
\usepackage{multirow}
\usepackage{diagbox}
\usepackage{indentfirst}
\usepackage{ulem}
\newtheorem{theorem}{Theorem}

\title{Localized LRR on Grassmann Manifolds: An Extrinsic View}

\begin{document}

\author{Boyue~Wang, 
        Yongli~Hu~\IEEEmembership{Member,~IEEE,} Junbin~Gao, Yanfeng~Sun~\IEEEmembership{Member,~IEEE,} and Baocai~Yin  ~\IEEEmembership{Member,~IEEE}
\thanks{Boyue Wang, Yongli Hu, and Yanfeng Sun are with Beijing Municipal Key Lab of Multimedia and Intelligent Software Technology, College of Metropolitan Transportation, Beijing University of Technology, Beijing 100124, China.
E-mail: boyue.wang@emails.bjut.edu.cn, \{huyongli,  yfsun\}@bjut.edu.cn}

\thanks{Junbin Gao is with the Discipline of Business Analytics, The University of Sydney Business School, The University of Sydney, NSW 2006, Australia. \protect E-mail: junbin.gao@sydney.edu.au}

\thanks{Baocai Yin is with the College of Computer Science and Technology, Faculty of Electronic Information and Electrical Engineering, Dalian University of Technology, Dalian 116620, China; and with Beijing Municipal Key Lab of Multimedia and Intelligent Software Technology at Beijing University of Technology, Beijing 100124, China. \protect E-mail: ybc@bjut.edu.cn
}}

\markboth{IEEE Transactions on Circuits and Systems for Video Technology,~Vol.~XX, No.~X, November~2016}%
{Wang \MakeLowercase{\textit{et al.}}: Kernelized LRR on Grassmann }

\maketitle

\begin{abstract}
Subspace data representation has recently become a common practice in many computer vision tasks. It demands generalizing classical machine learning algorithms for subspace data. Low-Rank Representation (LRR) is one of the most successful models for clustering vectorial data according to their subspace structures. This paper explores the possibility of extending LRR for subspace data on Grassmann manifolds.  Rather than directly embedding the Grassmann manifolds into the symmetric matrix space, an extrinsic view is taken to build the LRR self-representation in the local area of the tangent space at each Grassmannian point, resulting in a localized LRR method on Grassmann manifolds. A novel algorithm for solving the proposed model is investigated and implemented. The performance of the new clustering algorithm is assessed through experiments on several real-world datasets including MNIST handwritten digits, ballet video clips, SKIG action clips, DynTex++ dataset and highway traffic video clips. The experimental results show the new method outperforms a number of state-of-the-art clustering methods.
\end{abstract}

\IEEEpeerreviewmaketitle

\section{Introduction}

In recent years, action recognition has attracted a lot of attention, and is usually closely related to other lines of research that analyze motion from images and videos. Since each video clip always contains a different number of frames, na\"{i}vely vectorizing video data normally produces very high dimensional vectors, and the dimension of generated vectors varies for different vidoes. For example, a video clip with 10 frames produces a vector whose dimension is different from that of the resulting vector given by a video clip with 20 frames. Thus data have to be aligned to a proper and equal dimension before they can be used in any recognition learning algorithms. To overcome the dimension issue caused by simple vectorization, researchers have employed different strategies, from local statistical feature descriptors such as LBP-TOP (local binary pattern -- three orthogonal planes) \cite{ZhaoPietikaeinen2007}, linear dynamic systems ARMA (Autoregressive-Moving-Average) \cite{TuragaVeeraraghavanSrivastavaChellappa2011} to the subspace representation of the Grassmann manifold \cite{CetingulVidal2009}.

\begin{figure}
    \begin{center}
    \includegraphics[width=0.45\textwidth]{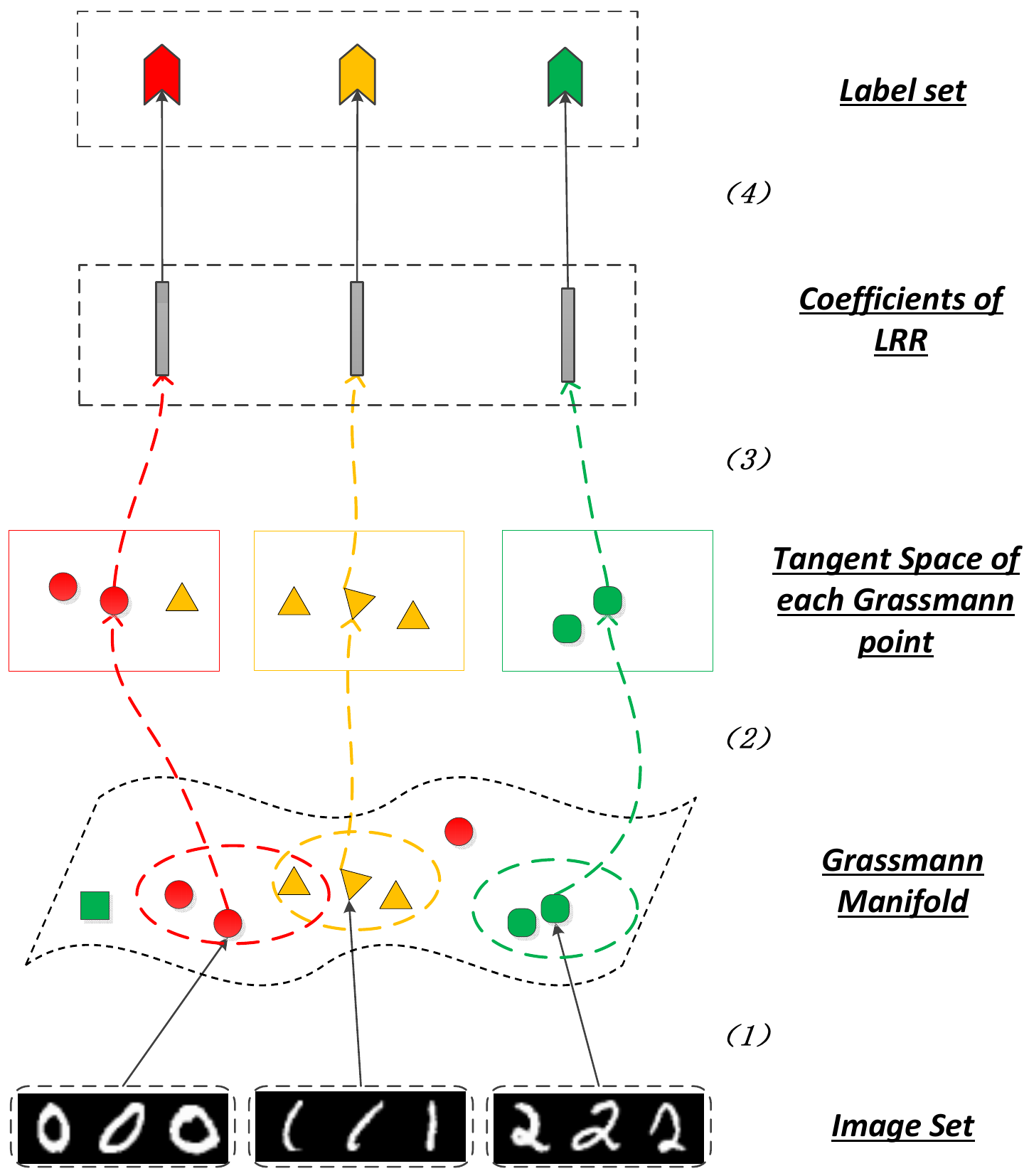}
    \end{center}
    \caption{(1) Image sets are represented as Grassmann points. (2) All Grassmann points are mapped onto the tangent space of each Grassmann point. (3) Build a linear representation in the tangent space of each Grassmann point to generate a holistic coefficient matrix by LRR. (4) Conduct clustering by NCuts.}\label{Fig1}
\end{figure}

Our main motivation here is to develop new methods for analyzing video data and/or image sets through subspace representations, by borrowing the ideas used in analyzing data samples in terms of linear spaces. It should be mentioned that the clustering problem researched in this paper is \emph{different from} the problem in conventional subspace clustering where the fundamental purpose is to cluster vectorial data samples in Euclidean spaces into clusters according to their subspace structures. That is, the objects to be clustered are vectorial samples in Euclidean space.  In this paper, the objects to be clustered  are themselves subspaces of the same dimension, i.e., the points on the abstract Grassmann manifold \cite{AbsilMahonySepulchre2004,EdelmanAriasSmith1998}, and the objective is to group these subspaces into several clusters according to certain criteria. For example, we may be interested in clustering a number of lines passing through the origin (one dimension subspaces) into several groups, where each group represents a rough direction. The relevant research can be seen in \cite{CetingulVidal2009}.

We distinguish \emph{clustering subspaces} (the subject of this paper) from the classical \emph{subspace clustering} (clustering vectorial data according to subspace structures).
Subspace clustering has attracted great interest in computer vision, pattern recognition, and signal processing \cite{ElhamifarVidal2013,WrightMaMairalSapiroHuang2010,XuWunsch-II2005,ZhangGhanemLiuAhuja2012,HeWangSuZhangLi2015,ChenDaiPanHuang2015}.  Vidal \cite{Vidal2011} classifies subspace clustering algorithms according to four categories: algebraic \cite{HongWrightHuangMa2006,Kanatani2001,MaYangDerksenFossum2008,ShiJiangMaoLuWang2015}, statistical \cite{GruberWeiss2004,TippingBishop1999a}, iterative \cite{HoYangLimLeeKriegman2003,Tseng2000,HeChenZhuWardCooperViantHeathYao2015}, and spectral clustering based approaches~\cite{ElhamifarVidal2013,WuHuGaoSunYin2015,LangLiuYuYan2012,LiuLinSunYuMa2013,TangLiuSuZhang2014,LuLiJiTanZhang2015}. The spectral clustering-based method consists of two steps: (1) learning a similarity matrix for the given data sample set; and (2) performing general clustering methods to categorize data samples such as Normalized Cuts (NCut) \cite{ShiMalik2000}.

Two  classical representatives of spectral clustering-based methods
are Sparse Subspace Clustering (SSC) \cite{ElhamifarVidal2013} and Low-Rank Representation (LRR) \cite{LiuLinSunYuMa2013}.
Both SSC and LRR rely on the self-expressive property in linear space \cite{ElhamifarVidal2013}: \textit{each data point in a union of subspace can be efficiently reconstructed by a linear combination of other points in the data set.}
SSC further induces sparsity by utilizing the $l_1$ Subspace Detection Property\cite{Donoho2004} in an independent manner, while the LRR model considers the intrinsic relation among the data objects in a holistic way via the low-rank requirement.
It has been proved that, when the data set is actually composed of a union of multiple subspaces, the LRR method can reveal this structure through subspace clustering \cite{LiuLinYu2010,ChenYang2014,XiaoTanXuDong2015,LiCuiDong2016,WongLaiXuWenHo2015}.

However, the principle of self-expression is only valid and applicable in classic linear/affine subspaces. As mentioned above, data samples on the Grassmann manifold are abstract. To apply the principle of self-expression on the Grassmann manifold, one has to concrete the abstract manifold and adopt ``linearity'' onto the manifold. It is well known \cite{Chikuse2002a} that the Grassmann manifold is isometrically equivalent to the symmetric idempotent
matrices subspace. To overcome the difficulty of SSC/LRR self-expression in the abstract Grassmann setting, the authors of \cite{WangHuGaoSunYin2014} embed the Grassmann manifold into the symmetric matrix manifold where the self-expression can be naturally defined in this embedding space, thus an LRR model on the Grassmann manifold was formulated. We refer to this method as the \textit{embedding} version. Embedding a manifold into a larger ambient Euclidean space is one of the approaches for dealing with tasks on manifolds.

The problem of clustering on manifolds has long been researched. Wang \emph{et al}. \cite{WangHuGaoSunYin2014} choose an embedding view to embed the image sets, the points on the Grassmann manifold, onto the symmetric matrix space  and extend the LRR model to the Grassmann manifold space (GLRR-F).
Turage \emph{et al}. \cite{TuragaVeeraraghavanSrivastavaChellappa2011} explore statistical modeling methods that are derived from the Riemannian geometry of the manifold (SCGSM).
Elhamifar \emph{et al}. \cite{ElhamifarVidalNips2011} investigate neighborhood characteristics around each data point and connect each point to its neighbors with appropriate weights for clustering purposes (SMCE).
Patel \emph{et al}. \cite{PatelNguyenVidal2013} propose a method that learns the projection of data and find sparse coefficients in the low-dimensional latent space (LS3C). All these manifold-based learning methods achieve great performance with regard to clustering problems, which are regarded as benchmark methods in this paper.

For our purpose, we start with a review on the geometric properties of the Grassmann manifold and take an extrinsic view to develop a new localized LRR or SSC method by exploring the local tangent space structure of the Grassmann manifold. The new formulation relies on so-called Log mapping on the Grassmann manifold through which the self-expression of Grassmann points can be realized in the local tangent space at each Grassmann point from the given dataset. This way, we extend the classical LRR method onto the Grassmann manifold and construct a localized LRR method. The local neighborhood is determined by geodesic distance on Grassmann manifold. Implementing linear self-expression in the local tangent space can be considered as the first-order approximation to the nonlinear relation on the manifold. Mapping a task on the manifold onto its tangent space is a standard extrinsic way to deal with problems in manifolds. Thus the proposed localized LRR on Grassmann manifold exploits the idea of using extrinsic linear relation to describe the non-linearity of high-dimensional data; thus clustering can ultimately be implemented on manifolds. In addition, compared with the classic LRR model working on single image data, our proposed method focus on clustering image set data.

The primary contributions of this paper are:
\begin{enumerate}
\item using an extrinsic view to formulate the LRR method on the Grassmann manifold by exploring the linear relation among all the lifted points in the tangent space at each Grassmann point. 
This proposed method has a totally different view with the conference paper FGLRR \cite{WangHuGaoSunYin2014} which is an embedding view to map Grassmann manifold onto Symmetrical matrix space;
\item incorporating a localization constraint into the standard LRR method to construct a localized LRR on Grassmann manifold; and
\item proposing a practical algorithm to solve the problem of
the localized LRR model on Grassmann manifold.
\end{enumerate}

The rest of the paper is organized as follows. In Section \ref{Sec:2}, we summarize the related works and the preliminaries for the Grassmann manifold. Section \ref{Sec:3} describes the low-rank representation on the Grassmann manifold. In Section \ref{Sec:4}, an algorithm for solving the LRR method on the Grassmann manifold is proposed. In Section \ref{Sec:5}, the performance of the proposed method is evaluated via the clustering problems of five
public datasets. Finally, conclusions and suggestions for future work are provided in Section \ref{Sec:6}.

\section{Preliminaries and Related Works}\label{Sec:2}

In this section, first, we briefly review the existing sparse subspace clustering methods, SSC and LRR. Then we describe the fundamentals of the Grassmann manifold that are relevant to our proposed method and algorithm.

\subsection{SSC \cite{ElhamifarVidal2013} and LRR \cite{LiuLinSunYuMa2013}}\label{Subsec:2.1}

Given a set of vectorial data $\mathbf{X} = [\mathbf x_1, \mathbf x_2, ..., \mathbf x_N]\in\mathbb{R}^{d\times N}$ drawn from an unknown union of subspaces, where $d$ is the data dimension and $N$ is the number of data samples, the objective of subspace clustering is to assign each data sample to its underlying subspace. The basic assumption is that the data in $\mathbf{X}$ are drawn from a union of $K$ subspaces $\{\mathcal{S}_k\}^K_{k=1}$ of dimensions $\{d_k\}^K_{k=1}$.

Under the data self-expressive principle, it is assumed that each point in the data set can be written as a linear combination of other points, i.e., $\mathbf{X} = \mathbf{X}\mathbf{Z}$, where $\mathbf{Z}\in\mathbb{R}^{N\times N}$ is a matrix of similarity coefficients. In the case of corrupted data, we may relax the self-expressive model to $\mathbf{X} = \mathbf{X}\mathbf{Z}+\mathbf E$, where $\mathbf E$ is the fitting error. With this model, the main purpose is to learn the expressive coefficients $\mathbf Z$ under some appropriate criteria, which can be used in a spectral clustering algorithm.

The optimization criterion for learning $\mathbf Z$ is
\begin{align}
\min_{\mathbf Z, \mathbf E}\|\mathbf E\|^2_\ell + \lambda \|\mathbf Z\|_q,  \text{ s.t. } \mathbf{X} = \mathbf{X}\mathbf{Z}+\mathbf E,\label{BasicSparseModel}
\end{align}
where $\ell$ and $q$ are the norm place-holders and $\lambda>0$ is a penalty parameter to balance the coefficient term and the reconstruction error.

The norm $\|\cdot\|_\ell$ is used to measure the data expressive errors while the norm $\|\cdot\|_q$ is chosen as a sparsity-inducing measure. The popular choice for $\|\cdot\|_\ell$ is Frobenius norm $\|\cdot\|_F$ (or $\|\cdot\|_2$). SSC and LRR differentiate at the choice of $\|\cdot\|_q$.  SSC is in favor of the sparsest representation for the data sample using $\ell_1$ norm \cite{Vidal2011} while LRR takes a holistic view in favor of a coefficient matrix in the lowest rank, measured by the nuclear norm $\|\cdot\|_*$ \cite{LiuLinSunYuMa2013}.

To avoid the case in which a data sample is represented by itself, an extra constraint $\text{diag}(\mathbf Z)=0$ is included in the SSC method. LRR uses the so-called $\ell_{2,1}$ norm to deal with random gross errors in data. In summary, we re-write the SSC method as follows,
\begin{align}
\min_{\mathbf Z, \mathbf E}\|\mathbf E\|^2_F + \lambda \|\mathbf Z\|_1,  \text{ s.t. } \mathbf{X} = \mathbf{X}\mathbf{Z}+\mathbf E, \text{diag}(\mathbf Z) = 0,\label{SSCModel}
\end{align}
and the LRR method
\begin{align}
\min_{\mathbf Z, \mathbf E}\|\mathbf E\|_{2,1} + \lambda \|\mathbf Z\|_*,  \text{ s.t. } \mathbf{X} = \mathbf{X}\mathbf{Z}+\mathbf E.\label{LRRModel}
\end{align}

In addition, the LRR model has been applied in multiple kind of manifolds. Wang \emph{et al}.\cite{WangHuGaoSunYin2014} employs an embedding view to embed Grassmann manifold to symmetrical matrix space to extend the classical LRR model. 
In previous researches, Yin \emph{et al}.\cite{YinGaoGuo2015}, Fu \emph{et al}.\cite{FuGaoHongTien2015} and Tierney \emph{et al}.\cite{TierneyGaoGuoZhang2016} implement the classic LRR model on Stiefel manifold, Symmetrical Positive Definite (SPD) manifold and Curves manifold respectively. Although these methods share the similar motivation with our proposed method that extends the LRR model onto manifolds, each manifold has different geometry and metric. In addition, these methods focus on single image data while our proposed method works on image set data.


\subsection{Grassmann Manifold}
This paper is particularly concerned with points on a known manifold. In most cases in computer vision applications, manifolds can be considered as low dimensional smooth ``surfaces'' embedded in a higher dimensional Euclidean space. A manifold is locally similar to Euclidean space around each point on the manifold.

In recent years, the Grassmann manifold \cite{BounmalAbsil2014,ConnieGohTeoh2016} has attracted great interest in the computer vision research community for numerous application tasks, such as subspace tracking \cite{SrivastavaKlassen2004}, clustering \cite{CetingulVidal2009}, discriminant analysis \cite{HammLee2008,XuWangGaoCaoTaoLiu2014}, and sparse coding \cite{HarandiSandersonShenLovell2013,HarandiSandersonShiraziLovell2011}. Mathematically the Grassmann manifold $\mathcal{G}(p,d)$ is defined as the set of all $p$-dimensional subspaces in $\mathbb{R}^d$, where $0\leq p\leq d$. Its Riemannian geometry has recently been well investigated in literature, e.g., \cite{AbsilMahonySepulchre2008,AbsilMahonySepulchre2004,EdelmanAriasSmith1998}.

As a point in $\mathcal{G}(p,d)$ is an abstract concept, a concrete representation must be chosen for numerical learning purposes. There are several classic concrete representations for the abstract Grassmann manifold in the literature. For our purpose, we briefly list some of  them. Denote by $\mathbb{R}^{d\times p}_*$ the space of all $d\times p$ matrices of full-column rank; $GL(p)$ the general group of nonsingular matrices of order $p$; $\mathcal{O}(p)$ the group of all the $p\times p$ orthogonal matrices; and $\mathcal{ST}(p,d) = \{\mathbf X\in\mathbb{R}^{d\times p}: \mathbf X^T\mathbf X = I_p\}$ the set of all the $p$ dimension bases, called the Stiefel manifold. Then four major matrix representations of the Grassmann manifold are

\textit{\uppercase\expandafter{\romannumeral1}. Representation by Full-Column Rank Matrices \cite{AbsilMahonySepulchre2008}:}
\[
\mathcal{G}(p,d) \cong \mathbb{R}^{d\times p}_* / GL(p).
\]

\textit{\uppercase\expandafter{\romannumeral2}. Orthogonal Representation \cite{EdelmanAriasSmith1998}:}
\[
\mathcal{G}(p,d) \cong \mathcal{O}(d) / \mathcal{O}(p) \times \mathcal{O}(d-p).
\]

\textit{\uppercase\expandafter{\romannumeral3}. The Stiefel Manifold Representation \cite{EdelmanAriasSmith1998}:}
\[
\mathcal{G}(p,d) \cong \mathcal{ST}(p,d) / \mathcal{O}(p).
\]

\textit{\uppercase\expandafter{\romannumeral4}. Symmetric Idempotent Matrix Representation \cite{Chikuse2002a}:}
\[
\mathcal{G}(p,d) \cong \{\mathbf{P}\in\mathbb{R}^{d\times d}:  \mathbf{P}^T = \mathbf{P}, \mathbf{P}^2=\mathbf{P}, \text{rank}(\mathbf{P}) = p\}.
\]

Many researchers adopt the symmetric idempotent matrix representation (IV) for learning tasks on the Grassmann manifold, for example, Grassmann Discriminant Analysis \cite{HammLee2008}, Grassmann Kernel Methods \cite{HarandiSalzmannJayasumanaHartleyLi2014}, Grassmann Low-Rank Representation \cite{WangHuGaoSunYin2014}, etc.

In representations I, II, and III, we identify a point $\mathbf X$ on the Grassmann manifold $\mathcal{G}(p,d)$ as an equivalent class under relevant quotient spaces. For example, in representation III, an abstract Grassmann point $\mathbf X\in\mathcal{G}(p,d)$ is realized as a representative $\mathbf{X}\in\mathcal{ST}(p,d)$ from the equivalent class $[\mathbf X] = \{\mathbf X\mathbf Q: \text{ for all } \mathbf Q\in\mathcal{O}(p)\}$. Two representatives $\mathbf X_1$ and $\mathbf X_2$ are equivalent if there exists an orthogonal matrix $\mathbf Q\in \mathcal{O}(p)$ such that $\mathbf X_1 = \mathbf X_2\mathbf Q$.

In this paper, we will work on the Stiefel Manifold Representation for the Grassmann manifold.  Each point on the Grassmann manifold is an equivalent class defined by
\begin{align}
[\mathbf X]  = \{\mathbf X\mathbf Q | \mathbf X^T\mathbf X = I_p, \mathbf Q \in\mathcal{O}(p)\}, \label{EquivalentClass}
\end{align}
where $\mathbf X$, a representative of the Grassmann point $[\mathbf X]$, is a point on the Stiefel manifold, i.e., a matrix of $\mathbb{R}^{d\times p}$ with orthogonal columns.

\subsection{Representing Image Sets and Videos on the Grassmann Manifold}
Our strategy is to represent the image sets or videos as subspace objects, i.e., points on the Grassmann manifold. For each image set or a video clip, we formulate the subspace by finding  a representative basis from the matrix of the raw features of image sets or videos with  SVD (Singular Value Decomposition),  as done in \cite{HarandiSandersonShenLovell2013,HarandiSandersonShiraziLovell2011}. Concretely, let $\{\mathbf Y_i\}_{i=1}^M$ be an image set, where $\mathbf Y_i$ is a grey-scale image with dimension $m \times n$ and $M$ is the number of all the images. For example, each $\mathbf Y_i$ in this set can be a handwritten digit 9 from the same person. We construct a matrix $\mathbf \Gamma=[\text{vec}(\mathbf Y_1), \text{vec}(\mathbf Y_2),..., \text{vec}(\mathbf Y_M)]$ of size $(m\times n)\times M$ by vectorizing raw image data $ \mathbf Y_i$. Then $\mathbf \Gamma$ is decomposed by SVD as $\mathbf \Gamma=\mathbf U \mathbf \Sigma \mathbf V$. We pick up the left $p$ columns of $\mathbf U$ as the Grassmannian point $[\mathbf X] = [\mathbf U(:,1:p)] \in \mathcal{G}(p,m\times n)$ to represent the image set $\{\mathbf{Y}_i\}^M_{i=1}$.

\section{LRR on the Grassmann Manifold}\label{Sec:3}
We develop the LRR model on the Grassmann manifold in this section.
\subsection{The Idea and the Model}
The famous \textit{Science} paper written by Roweis and Saul \cite{RoweisSaul2000} proposes a manifold learning algorithm for low-dimension space embedding by approximating the manifold structure of data. The method is named Locally Linear Embedding (LLE). Similarly a latent variable version, Local Tangent Space Alignment (LTSA), has been proposed in \cite{ZhangZha2004}.

Both LLE and LTSA exploit the ``manifold structures'' implied by data. In this setting, we have no knowledge of the manifold where the data reside, and the manifold structure is to be learned from the data. This type of tasks is called \textit{manifold learning}. However, in many computer vision tasks, we explicitly know the manifold where the data come from. The idea of using manifold information to assist in learning tasks can be seen in earlier research work \cite{LeeAbbottAraman2009} and recent work \cite{HarandiSalzmannHartley2014}. This type of task falls in the new paradigm of \textit{learning on manifolds}.

To construct the LRR method on manifolds, we first review the LLE algorithm. LLE relies on learning the local linear combination $\mathbf X_i \approx \sum\limits_{j=1}^{C} w_{ii_j}\mathbf X_{i_j}$, where $C$ is the cardinality of a neighborhood $\mathcal{N}_i = \{\mathbf X_{i_j}\}^C_{j=1}$ of $\mathbf X_i$ (including $\mathbf X_i$ such that $\mathbf X_{i_1} = \mathbf X_i$) under the condition $\sum\limits_{j=1}^{C} w_{ii_j} = 1$ (assume $w_{ii_1}=0$ corresponding to $\mathbf X_i$). Hence
\begin{align}
\sum_{j=1}^{C} w_{ii_j}\mathbf X_{i_j} - \mathbf X_i = \sum\limits_{j=1}^{C}w_{ii_j}(\mathbf X_{i_j} - \mathbf X_i) \approx 0. \label{TangentApproximation}
\end{align}

Define $w_{ij}=0$ for any $\mathbf X_j\not\in\mathcal{N}_i$, then \eqref{TangentApproximation}  can be written as  a specific form of the data self-expression in \eqref{BasicSparseModel}
\begin{align}
\begin{aligned}
&\sum\limits_{j=1}^{N}w_{ij}(\mathbf X_{j} - \mathbf X_i) \approx 0,  \;\text{s.t.}\; \sum^N_{j=1} w_{ij} = 1.\label{TangentApproximation1}
\end{aligned}
\end{align}

Under the manifold terms, the vector $\mathbf X_{j}-\mathbf X_i$ can be regarded as a tangent vector at point $\mathbf X_i$ under the plain Euclidean manifold.  Thus \eqref{TangentApproximation1} means that the linear combination of all these tangent vectors should be close to the 0 tangent vector. The tangent vector $\mathbf X_{j}-\mathbf X_i$ at $\mathbf X_i$ under Euclidean space can be extended to the Log mapping $\text{Log}_{\mathbf X_i}(\mathbf X_{j})$ on general manifolds. The Log mapping maps $\mathbf X_{j}$ on the manifold to a tangent vector at $\mathbf X_i$. As a result, the data self-expressive principle used in \eqref{TangentApproximation1} can be realized on the tangent space at each data $\mathbf X_i$ on a general manifold as
\begin{align}
\begin{aligned}
\sum\limits_{j=1}^{N} w_{ij}\text{Log}_{\mathbf X_i}(\mathbf X_{j}) \approx 0, \;
\text{ s.t }\; \sum_{j=1}^N w_{ij}=1, \label{TangentApproximation2}
\end{aligned}
\end{align}
where we assume that $w_{ij} = 0$ if $\mathbf X_j \not\in\mathcal{N}_i$ and specially $w_{ii}=0$.

In fact, this idea of lifting the linear relation over to tangent space has been used in \cite{CetingulWrightThompsonVidal2014,GohVidal2008}. Goh \emph{et al}. \cite{GohVidal2008} extend nonlinear dimensionality reduction (i.e., Laplacian Eigenmaps, Locally Linear Embedding and Hessian LLE) on the SPD manifold to deal with motion segmentation and DTI segmentation; and  Cetingul \emph{et al}. \cite{CetingulWrightThompsonVidal2014} use sparse representations to segment high angular resolution diffusion imaging data described by Orientation Distribution Function. Different from \cite{CetingulWrightThompsonVidal2014,GohVidal2008}, we adopt an extended LRR method for clustering on Grassmann manifold. Additionally, \cite{CetingulWrightThompsonVidal2014,GohVidal2008} handle the single image data while our proposed method clusters image set data.

The above request about the linear relation on tangent space was also obtained in \cite{XieHoVemuri2013} by approximately defining a linear combination on the manifold for the purpose of dictionary learning over Riemannian manifold. It is also pointed out in \cite{XieHoVemuri2013} that the affine constraints $\sum^N_{j=1}w_{ij} = 1$ ($i=1, 2, ..., N$) can preserve the coordinate independence on manifold. 

Finally  the LRR/SSC on the manifold is formulated as the following optimization problem:
\begin{align}
\begin{aligned}
&\min_{\mathbf W}\frac12\sum^N_{i=1}\bigg\|\sum^N_{j=1}w_{ij}\text{Log}_{\mathbf X_i}(\mathbf X_j)\bigg\|^2_{X_i} + \lambda \|\mathbf W\|_q  \\
&\text{s.t.} \sum^N_{j=1}w_{ij} = 1,  i=1,2,..., N. \text{ and } P_{\mathbf \Omega}(\mathbf W)=0,
\end{aligned}\label{25August2014-3}
\end{align}
where $P_{\mathbf \Omega}(\mathbf W) = 0$ is a constraint to preserve the formal local properties of the coefficients, so the coefficient matrix $\mathbf W$ is sparse. Here $P_{\mathbf \Omega}(\mathbf W) = 0$ is implemented by a projection operator $P_{\mathbf \Omega}$ over the entries of $\mathbf W$ defined as follows:
\begin{equation}\label{Project}
    P_{\mathbf \Omega}(w_{i,j})=
    \begin{cases}
    0 & \text{if } (i,j)\in \mathbf \Omega\\
    w_{ij} & \text{otherwise},
    \end{cases}
\end{equation}
where the index set of $\Omega$ is defined as
\[
\mathbf \Omega = \{(i,j): j=i \text{ or } \mathbf X_j\notin \mathcal{N}_i\}.
\]
To preserve the local properties of the coefficients, there are a number of ways to pre-define the index set $\mathbf  \Omega$. For example, we can use a threshold over the Euclidean distance between $\mathbf X_i$ and $\mathbf X_j$ in the ambient space of the manifold, or we may use a threshold over the geodesic distance between $\mathbf X_i$ and $\mathbf X_j$. In this paper, we adopt the KNN (K-nearest neighbor) strategy to choose $C$ closest neighbors under geodesic distance. Thus the neighborhood size $C$ is a tunable parameter in our method.

When $\|\cdot\|_q$ in \eqref{25August2014-3} is $\|\cdot\|_1$, we have the SSC method on manifolds, and when $\|\cdot\|_q$ takes $\|\cdot\|_*$, the LRR on the manifold is formed. Part of this idea has been used in the LRR method on the Stiefel manifold~\cite{YinGaoGuo2015}, the Curves manifold~\cite{TierneyGaoGuoZhang2016} and the symmetric positive definite matrix (SPD) manifold~\cite{FuGaoHongTien2015}. In the sequel, we mainly focus on LRR on the Grassmann manifold generated from the above tangent space operation. It is also worthwhile pointing out that the new model is conceptually different from both SSC and LRR due to incorporating neighhorhood information. Since the coefficient matrix $\mathbf W$ is sparse and low-rank, the proposed method is called Localized LRR on Grassmann manifold (LGLRR).

\subsection{Properties of the Grassmann Manifold Related to LRR Learning}
For this special Grassmann manifold, formulation \eqref{25August2014-3} is actually abstract. The objects we are working with are equivalence classes $[\mathbf X]$. We need to specify the meaning of the norm $\|\cdot\|_{[\mathbf X]}$ and $\text{Log}_{[\mathbf X]}([\mathbf Y])$.
\subsubsection{Tangent Space, Its Metric, and Geodesic Distance}
Consider a point $\mathcal{S} = \text{span}(\mathbf X)$ on the Grassmann manifold $\mathcal{G}(p,d)$ where $\mathbf X$ is a representative of the equivalent class $[\mathbf X]$ defined in \eqref{EquivalentClass}. We denote the tangent space at $\mathcal{S}$ by $T_{\mathcal{S}}\mathcal{G}(p,d)$.  A tangent vector $\mathcal{H}\in T_{\mathcal{S}}\mathcal{G}(p,d)$ is represented by a matrix $\mathbf H\in\mathbb{R}^{d\times p}$ verifying, see \cite{BounmalAbsil2014},
\[
\frac{d}{dt}\text{span}(\mathbf X+t\mathbf H)|_{t=0} = \mathcal{H}.
\]

Under the condition that $\mathbf X^T\mathbf H=0$, it can be proved that $\mathbf H$ is the unique matrix as the representation of tangent vector $\mathcal{H}$, known as its horizontal lift at the representative $\mathbf X$. Hence the abstract tangent space  $T_{\mathcal{S}}\mathcal{G}(p,d)$ at $\mathcal{S} = \text{span}(\mathbf X)$ (or equivalently the equivalent class $[\mathbf X]$ of $\mathbf X$) can be represented by the following concrete set
\[
T_{[\mathbf X]}\mathcal{G}(p,d) = \{\mathbf H\in\mathbb{R}^{d\times p} |  \mathbf X^T\mathbf H = 0\}.
\]

The above representative tangent space is embedded in matrix space $\mathbb{R}^{d\times p}$ and it inherits the canonical inner
\begin{align*}
\forall \mathbf H_1, \mathbf H_2\in T_{[\mathbf X]}\mathcal{G}(p,d), \ \langle \mathbf H_1, \mathbf H_2\rangle_{[\mathbf X]} = \text{trace}(\mathbf H^T_1\mathbf H_2).  
\end{align*}
Under this metric, the Grassmann manifold is Riemannian and the norm term used in \eqref{25August2014-3} becomes
\begin{align}
\| \mathbf H \|^2_{[\mathbf X]}= \langle \mathbf H, \mathbf H\rangle_{[\mathbf X]} = \text{trace}(\mathbf H^T\mathbf H)
\label{Norm}
\end{align}
which is irrelevant to the point $[\mathbf X]$.  Note that $\mathbf H$ is a representative of a tangent vector at $[\mathbf X]$ on the Grassmann manifold.

The geodesic on the Grassmann manifold has been explored in~\cite{EdelmanAriasSmith1998}. For our purpose of finding the $C$ closest neighbors of a Grassmann point, we review the geodesic distance for any two Grassmann points here. Suppose both $[\mathbf X], [\mathbf Y]\in\mathcal{G}(p,d)$ with $\mathbf X, \mathbf Y\in\mathcal{ST}(p,d)$ as representatives, respectively. Take the SVD of $\mathbf X^T\mathbf Y$ as $\mathbf U\mathbf S\mathbf V = \mathbf X^T\mathbf Y$ such that $\mathbf S = \text{diag}(\mathbf s_i)$, then the geodesic distance between Grassmann points $\mathbf X$ and $\mathbf Y$ is defined as the summation of squared principal angles,
\begin{align}
d(\mathbf X, \mathbf Y)^2 = \sum_{i} \arccos^2(\mathbf s_i).\label{distance2}
\end{align}

\subsubsection{Log Mapping of the Grassmann Manifold} \label{ExpLog}
Log map on the general Riemannian manifold is the inverse of the Exp map on the manifold \cite{Lee2002}. For the Grassmann manifold, there is no explicit expression for the Log mapping. However, the Log operation  can be written out by the following algorithm \cite{EdelmanAriasSmith1998,RentmeestersAbsilDooren2010}:

Given two representative Stiefel matrices $\mathbf X, \mathbf Y$ of equivalent classes $[\mathbf X], [\mathbf Y]\in\mathcal{G}(p, d)$ as points on the Grassmann manifold (matrix representations in $d$-by-$p$ matrices), we seek to find $\mathbf H\in T_{[\mathbf X]}\mathcal{G}(p,d)$  such that the exponential map $\text{Exp}_{[\mathbf X]} (\mathbf H) = \mathbf Y$.

Instead of $\mathbf H$, we equivalently identify its thin-SVD $\mathbf H = \mathbf U\mathbf S\mathbf V^T$. Let us consider these equations, where $\mathbf Y$ is obtained with the exponential map $\text{Exp}_{[\mathbf X]} (\mathbf U\mathbf S\mathbf V^T)$,
\[
\mathbf Y = \mathbf X \mathbf V \cos(\mathbf S) \mathbf Q^T + \mathbf U \sin(\mathbf S) \mathbf Q^T,
\]
where $\mathbf Q$ is  any $p$-by-$p$  orthogonal matrix. For simplification, one choice is $\mathbf Q=\mathbf V$.

Consequently, we need to solve the equations of $\mathbf U, \mathbf S, \mathbf V$, and $\mathbf Q$, with the knowledge that $\mathbf X^T \mathbf H = 0$, since $\mathbf H$ is in the tangent space,
\begin{align*}
\mathbf V \cos(\mathbf S) \mathbf Q^T = \mathbf X^T\mathbf Y, \\
\mathbf U \sin(\mathbf S) \mathbf Q^T = \mathbf Y - \mathbf X \mathbf X^T\mathbf Y.
\end{align*}
Hence
\begin{align}
\mathbf U \sin(\mathbf S) \mathbf Q^T (\mathbf V\cos(\mathbf S)\mathbf Q^T)^{-1} = (\mathbf Y - \mathbf X \mathbf X^T\mathbf Y)(\mathbf X^T\mathbf Y)^{-1}
\label{LogGrassmannInverse}
\end{align}
which gives
\[
\mathbf U \tan(\mathbf S) \mathbf V^T  = (\mathbf Y - \mathbf X \mathbf X^T\mathbf Y)(\mathbf X^T\mathbf Y)^{-1}
\]
because $\mathbf V$ is actually an orthogonal matrix.  Hence the algorithm will conduct a SVD decomposition of $\mathbf U\Sigma \mathbf V^T = (\mathbf Y - \mathbf X \mathbf X^T\mathbf Y)(\mathbf X^T\mathbf Y)^{-1}$, then define
\begin{align}
\text{Log}_{[\mathbf X]}([\mathbf Y]) = \mathbf H = \mathbf U \arctan(\Sigma) \mathbf V^T.
\label{LogGrassmann}
\end{align}

It should be noted here that $\mathbf H$ does not always have full rank. Thus, the operation $(\mathbf V \cos(\mathbf S) \mathbf Q^T)^{-1}$ in \eqref{LogGrassmannInverse} is not always computable. Fortunately, it is not necessary to compute this inverse in the final Log mapping \eqref{LogGrassmann}.

\section{Solving LRR on the Grassmann Manifold}\label{Sec:4}
\subsection{Optimization for LGLRR}
With the concrete specifications \eqref{Norm} and \eqref{LogGrassmann} for the Grassmann manifold, the problem for LRR on the Grassmann manifold \eqref{25August2014-3} becomes numerically computable.
Denote by
\begin{align}
B^i_{jk} = \text{trace}(\text{Log}_{[\mathbf X_i]}([\mathbf X_j])^T \text{Log}_{[\mathbf X_i]}([\mathbf X_k])), \label{B}
\end{align}
which can be calculated according to the Log algorithm as defined in \eqref{LogGrassmann} for all the given data $\mathbf X_i$'s. Then problem \eqref{25August2014-3} can be rewritten as
\begin{align}
\begin{aligned}
&\min_{W}\frac12\sum^N_{i=1} \mathbf w_i \mathbf B_i \mathbf w^T_i + \lambda \|\mathbf W\|_*  \\
&\text{s.t.}\;\; \mathbf W\mathbf{1} = \mathbf{1}, \ \ P_{\mathbf \Omega}(\mathbf W)=0;
\end{aligned}\label{25August2014-4}
\end{align}
where $\mathbf B_i = (B^i_{jk})$, $\mathbf w_i = (w_{i1}, ..., w_{iN})$ is the row vector of $\mathbf W$ and $\mathbf 1\in \mathbb{R}^N$ is the column vector of all ones.

The augmented Lagrangian objective function of \eqref{25August2014-4} is given by
\begin{equation}
\begin{aligned}
L = &\lambda \|\mathbf W\|_*  + \frac12\sum^N_{i=1}\mathbf w_i \mathbf B_i\mathbf w^T_i + \langle \mathbf {Y}_1, \mathbf W\mathbf 1 - \mathbf 1\rangle \\
&+ \langle \mathbf Y_2, \mathbf W\rangle_{\mathbf \Omega} + \frac{\beta}2\left( \|\mathbf W\mathbf 1 - \mathbf 1\|^2_F + \|\mathbf W\|^2_{F,\mathbf \Omega} \right)
\end{aligned}\label{25August2014-5}
\end{equation}
where $\mathbf Y_1$ is the Lagrangian multiplier (vector) corresponding to the equality constraint $\mathbf W\mathbf 1 = \mathbf 1$, $\mathbf Y_2$ is the Lagrangian multiplier (matrix in the same size as $\mathbf W$) with dummy elements outside the positions $\mathbf \Omega$, 
and $\|\cdot\|_{F,\mathbf \Omega}$ is the F-norm defined over the elements over $\mathbf \Omega$. We will use  an adaptive way to update the constant $\beta$ in the iterative algorithm to be introduced.

Denote by $F(\mathbf W)$ the function defined by \eqref{25August2014-5} except for the first term $\lambda \|\mathbf W\|_*$. To solve \eqref{25August2014-5}, we adopt a linearization of $F(\mathbf W)$ at the current location $\mathbf W^{(k)}$ in the iteration process, that is, we approximate $F(\mathbf W)$ using the following linearization with a proximal term
\begin{align*}
F(\mathbf W)\approx & F(\mathbf W^{(k)}) + \langle \partial F(\mathbf W^{(k)}), \mathbf W- \mathbf W^{(k)}\rangle \\
&+ \frac{\eta_W\beta_k}2\|\mathbf W- \mathbf W^{(k)}\|^2_F,
\end{align*}
where $\eta_W$ is an approximate constant with a suggested value given by $\eta_W = \max\{\|\mathbf B_i\|^2\}+N+1$, and $\partial F(\mathbf W^{(k)})$ is the gradient matrix of $F(\mathbf W)$ at $\mathbf W^{(k)}$. Denote by $\mathcal{B}$ the 3-order tensor whose $i$-th front slice is given by $\mathbf B_i$. Let us define $\mathbf W\odot \mathcal{B}$ the matrix whose $i$-th row is given by $\mathbf w_i \mathbf B_i$, then it is easy to show
\begin{equation}
\begin{aligned}
\partial F(\mathbf W^{(k)}) = & \mathbf W\odot \mathcal{B} + \mathbf Y_1\mathbf 1^T + \beta^{(k)} (\mathbf W\mathbf 1 - \mathbf 1)\mathbf 1^T \\
& + P_{\Omega}(\mathbf Y_2) + \beta^{(k)} P_{\Omega}(\mathbf W^{(k)}).
\end{aligned}\label{Eq:14October2014-4}
\end{equation}

Hence \eqref{25August2014-5} can be approximately solved using the following iteration
\begin{align}
 \mathbf W^{(k+1)}   
= &\arg\min_W \lambda \|\mathbf W\|_* \label{SolutionW}\\
+&\frac{\eta_W\beta^{(k)}}{2} \bigg\|\mathbf W - \left(\mathbf W^{(k)} - \frac{1}{\eta_W\beta^{(k)}} \partial F(\mathbf W^{(k)})\right)\bigg\|^2_F. \notag
\end{align}

Problem \eqref{SolutionW} admits a closed-form solution by using the SVD thresholding operator \cite{CaiCandesShen2008}, given by
\begin{equation}\label{SolutionWk}
\begin{aligned}
\mathbf W^{(k+1)} = \mathbf U_{W} S_{\frac{\lambda}{\eta_W\beta^{(k)}}}(\mathbf \Sigma_{W})\mathbf V_{W}^{T},
\end{aligned}
\end{equation}
where $\mathbf U_{W} \mathbf \Sigma_{W} \mathbf V_{W}^{T}$ is the SVD of $\mathbf W^{(k)} - \frac{1}{\eta_W\beta^{(k)}} \partial F(\mathbf  W^{(k)})$ and $S_\tau(\cdot)$ is the Singular Value Thresholding (SVT)  operator \cite{CaiCandesShen2008} defined by
\begin{equation}
\begin{aligned}
S_{\tau}(\Sigma) = \text{diag}(\text{sgn}(\Sigma_{ii})(|\Sigma_{ii}|-\tau)).
\end{aligned}
\end{equation}

The updating rules for $\mathbf Y_1$ and $\mathbf Y_2$ are
\begin{equation}
\begin{aligned}
\mathbf Y^{(k+1)}_1 = \mathbf Y^{(k)}_1 + \beta^{(k)} (\mathbf W^{(k)}\mathbf 1 -\mathbf 1), \\
P_{\mathbf \Omega}(\mathbf Y^{(k+1)}_2) = P_{\mathbf \Omega}(\mathbf Y^{(k)}_2) + \mu^{(k)} P_{\mathbf \Omega}(\mathbf W^{(k)}),
\end{aligned}\label{Eq:14October2014-5}
\end{equation}
and the updating rule for $\beta_k$ is
\begin{equation}
\begin{aligned}
\beta^{(k+1)} = \min\{\beta_{\text{max}}, \rho \beta^{(k)}\},
\end{aligned}\label{UpdateBeta}
\end{equation}
where
 \[
 \rho = \begin{cases} \rho_0 & \beta^{(k)} \|\mathbf W^{(k+1)} - \mathbf W^{(k)}\| \leq \varepsilon_1,\\
 1 & \text{otherwise}.
 \end{cases}
 \]

Finally, the convergence condition can be defined as,
\begin{equation}\label{convergence}
\begin{aligned}
\beta^{(k)} \|\mathbf W^{(k+1)} - \mathbf W^{(k)}\| \leq \varepsilon_1 \ \ \text{and} \ \ \|\mathbf W^{(k+1)}\mathbf 1 - \mathbf 1\| \leq \varepsilon_2
\end{aligned}
\end{equation}

We summarize the above as Algorithm~\ref{wholeAlg1a}.

Once the coefficient matrix $\mathbf W$ is found, a spectral clustering like NCut \cite{ShiMalik2000} can be applied on the affinity matrix $\frac{|\mathbf W|+|\mathbf W|^T}{2}$ to obtain segmentation of the data.

\begin{algorithm}
\renewcommand{\algorithmicrequire}{\textbf{Input:}}
\renewcommand\algorithmicensure {\textbf{Output:} }
\caption{ Localized LRR on Grassmann Manifold.}\label{wholeAlg1a}
\begin{algorithmic}[1]
\REQUIRE The Grassmann sample set $\{\mathbf X_i\}_{i=1}^N$,$\mathbf X_i\in \mathcal{G}(p,d)$, the projected matrix $\Omega$ and the balancing parameter $\lambda$.  \\
\ENSURE  The Low-Rank Representation $\mathbf W$ ~~\\
\STATE   Initialize: Set the parameters $\rho_0=1.9$, $\eta_W = \max\{\|\mathbf B_i\|^2\}+N+1$, $\beta_{\max}=10^6\gg \beta_0 = 0.1$, $\varepsilon_1 = 10e-4$, $\varepsilon_2 = 10e-4$, $\mathbf W^{(0)} = 0$, $\mathbf Y^{(0)}_1=0$, $\mathbf Y^{(0)}_2=0$. \\
\STATE   Prepare all $\mathbf B_i$s according to \eqref{B}
\WHILE   {not converged}
\STATE   Calculate $\partial F(\mathbf W^{(k)})$  according to \eqref{Eq:14October2014-4}
\STATE   Let $\mathbf M^{(k)} = \mathbf W^{(k)} -  \frac1{\eta_B\beta_k} \partial F(\mathbf W^{(k)})$, then
\STATE   Update the $\mathbf W^{(k+1)}$ according to \eqref{SolutionW}
\STATE   Update $\mathbf Y^{(k+1)}_1$ and $\mathbf Y^{(k+1)}_2$ by \eqref{Eq:14October2014-5}
\STATE   Update $\beta^{(k+1)}$  by \eqref{UpdateBeta}
\STATE   Check the convergence condition by \eqref{convergence}
\ENDWHILE
\end{algorithmic}
\end{algorithm}

\subsection{Complexity Analysis}
For ease of analysis, we firstly define some symbols used in the following discussion. Let $K$ and $r$ denote the total number of iterations and the lowest rank of the matrix $\mathbf W$, respectively. The size of $\mathbf W$ is $N\times N$. The major computation cost of our proposed method contains two parts, calculating all the $\mathbf B_i$'s and updating $\mathbf W$. In terms of formula \eqref{LogGrassmann}, the computational complexity of Log algorithm is $O(p^3)$ thanks to a SVD decomposition over a matrix of size $d\times p$. Trace-norm's cost in \eqref{B} is $O(d^2)$; therefore, the complexity of $B_{jk}^i$ is at most $O(d^3)$ and $\mathbf B_i$'s computational complexity is $O(N^2d^3)$. Thus the total for all the $\mathbf B_i$ is $O(N^3d^3)$. In each iteration of the algorithm, singular value thresholding is adopted to update the low-rank matrix $\mathbf W$ whose complexity is $O(rN^2)$~\cite{LiuLinSunYuMa2013}. Suppose the algorithm is terminated after $K$ iterations, then the overall computational complexity is given by
\[
O(N^3d^3)+O(KrN^2).
\]

\subsection{Convergence Analysis}
Algorithm~\ref{wholeAlg1a} is adopted from the algorithm proposed in~\cite{LinLiuSu2011}. However, owing to the terms of $\mathbf B_i$'s in the objective function~\eqref{25August2014-5}, the convergence theorem proved in~\cite{LinLiuSu2011} cannot be directly applied to this case as the linearization is implemented on both the augmented Lagrangian terms and the term involving $\mathbf B_i$'s. Fortunately we can employ the revised approach, presented in the new paper~\cite{YinGaoLinShiGuo2015}, to prove the convergence for the algorithm. Without repeating all the details, we present the convergence theorem for Algorithm~\ref{wholeAlg1a} as follows,
\begin{theorem}[Convergence of Algorithm~\ref{wholeAlg1a}] If $\eta_W\geq \max\{\|\mathbf B_i\|^2\}+N+1$, $\displaystyle\sum^{+\infty}_{k=1}\beta^{{(k)}^{-1}} = +\infty$, $\displaystyle\beta^{(k+1)}-\beta^{(k)} > C_0 \frac{\sum_i \|\mathbf B_i\|^2}{\eta_W - \max\{\|\mathbf B_i\|^2\}-N}$, where $C_0$ is a given constant and $\|\cdot\|$ is the matrix spectral norm, then the sequence $\{\mathbf W^{k}\}$ generated by Algorithm~\ref{wholeAlg1a} converges to an optimal solution to problem~\eqref{25August2014-4}.
\end{theorem}

\begin{figure}[!h]
    \begin{center}
    \includegraphics[width=0.45\textwidth]{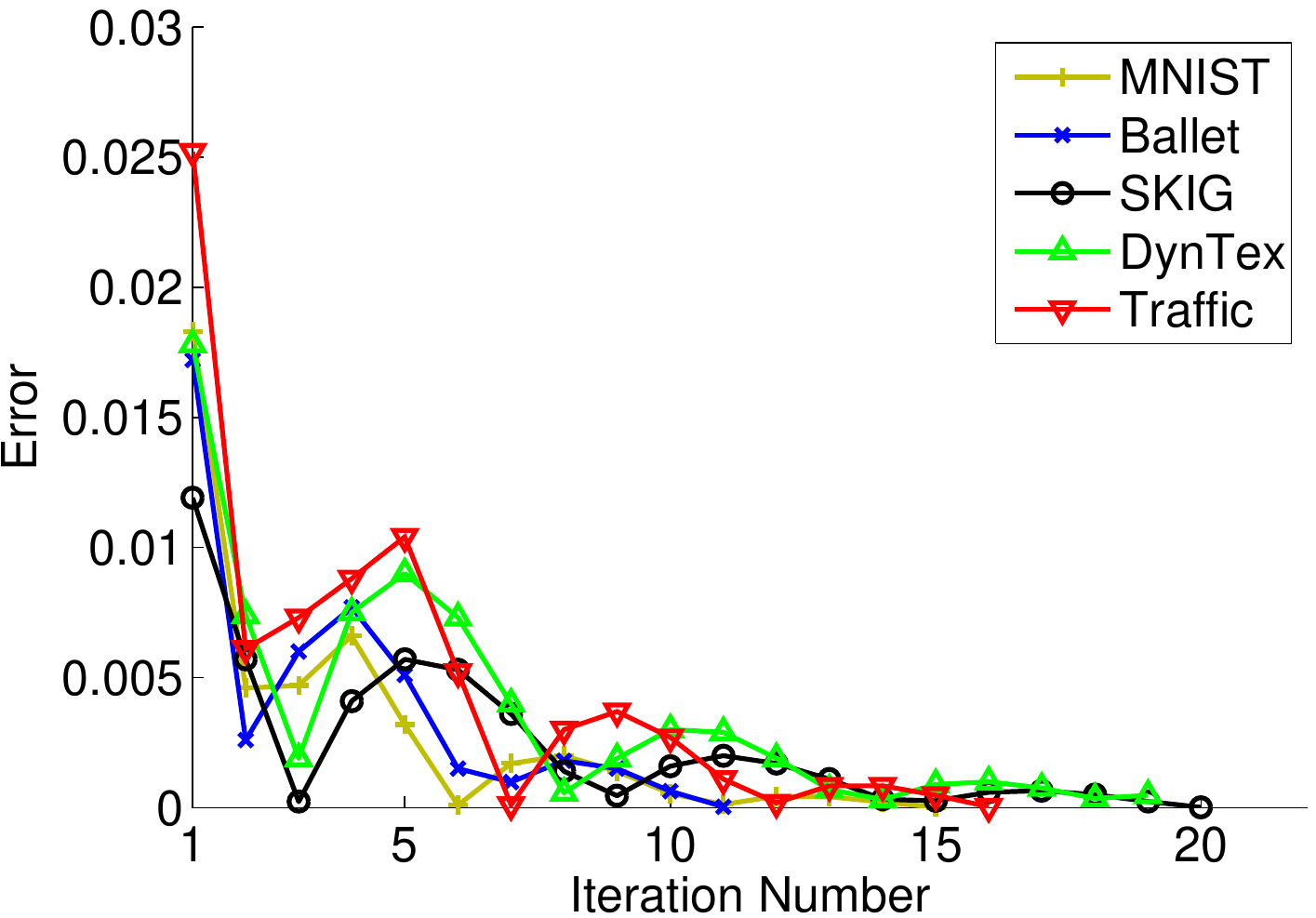}
    \end{center}
    \caption{Convergence behavior of our proposed methods on all datasets. The error value in each iteration equals to $\max\{\beta^{(k)} \|\mathbf W^{(k+1)} - \mathbf W^{(k)}\|  , \|\mathbf W^{(k+1)}\mathbf 1 - \mathbf 1\| \}$}\label{FigConvergence}
\end{figure}

We give an intuitive illustration of the convergence with respect to the iteration number on all datasets, as shown in Fig. \ref{FigConvergence}. It is demonstrated that our optimization algorithm has a strong convergence property. For all datasets that we experimented, the algorithm can be converged in about 20 iteration steps.

\section{Experiments}\label{Sec:5}

For convenience, we refer to the proposed model ~\eqref{25August2014-4} as the Localized Grassmann Low-Rank Representation model (LGLRR).
In this section, we compare the performance of the proposed LGLRR method against the benchmark spectral clustering methods, Low-Rank Representation on Grassmann Manifold (GLRR-F) \cite{WangHuGaoSunYin2014}, Low-Rank Representation (LRR) \cite{LiuLinSunYuMa2013} and Sparse Subspace Clustering (SSC) \cite{ElhamifarVidal2013}, and four manifold learning clustering methods, Low Rank Representation on Stiefel manifolds (SLRR) \cite{YinGaoGuo2015}, Statistical computations on Grassmann and Stiefel manifolds (SCGSM) \cite{TuragaVeeraraghavanSrivastavaChellappa2011}, Sparse Manifold Clustering and Embedding (SMCE) \cite{ElhamifarVidalNips2011}, and Latent Space Sparse Subspace Clustering (LS3C) \cite{PatelNguyenVidal2013}.

We evaluate the performances on the clustering task for five widely used datasets: MNIST Handwritten dataset\footnote{\url{http://yann.lecun.com/exdb/mnist/}.}, ballet video clips\footnote{\url{https://www.cs.sfu.ca/research/groups/VML/semilatent/}.}, SKIG action clips\footnote{\url{http://lshao.staff.shef.ac.uk/data/SheffieldKinectGesture.htm}.}, DynTex++ dataset\footnote{\url{http://projects.cwi.nl/dyntex/database.html}.}, and Highway Traffic clips\footnote{\url{http://www.svcl.ucsd.edu/projects/traffic/}.}.

\subsection{Experimental Preparation}
\subsubsection{Setting Model Parameters}
We acknowledge that some parameters should be adequately adjusted in the proposed method, such as $\lambda$, $\varepsilon$, and $C$.

$\lambda$ is the most important penalty parameter for balancing the error term and the low-rank term in our proposed method. Empirically, the best value of $\lambda$ depends on  the application problems and has to be chosen from a large range of values to get a better performance in a particular application. From our experiments, we have observed that when the cluster number increases, the best $\lambda$ decreases. Additionally, $\lambda$ will be smaller when the noise level in the data is lower while $\lambda$ will become larger if the noise level is higher. These observations are useful in selecting a proper $\lambda$ value for different datasets.

The parameter $C$ defines the neighborhood size of each Grassmann point. From the experiments, we have noted that an appropriate size $C$ depends on both the number of samples in different clusters and the distribution of points on the Grassmann manifold. Our empirical studies suggest that a size slightly larger than the average value of the number of data in different clusters provides satisfactory results.

\begin{figure*}
\centering
\subfigure[]{ \label{Orderfig:a} 
\includegraphics[width=0.3\textwidth]{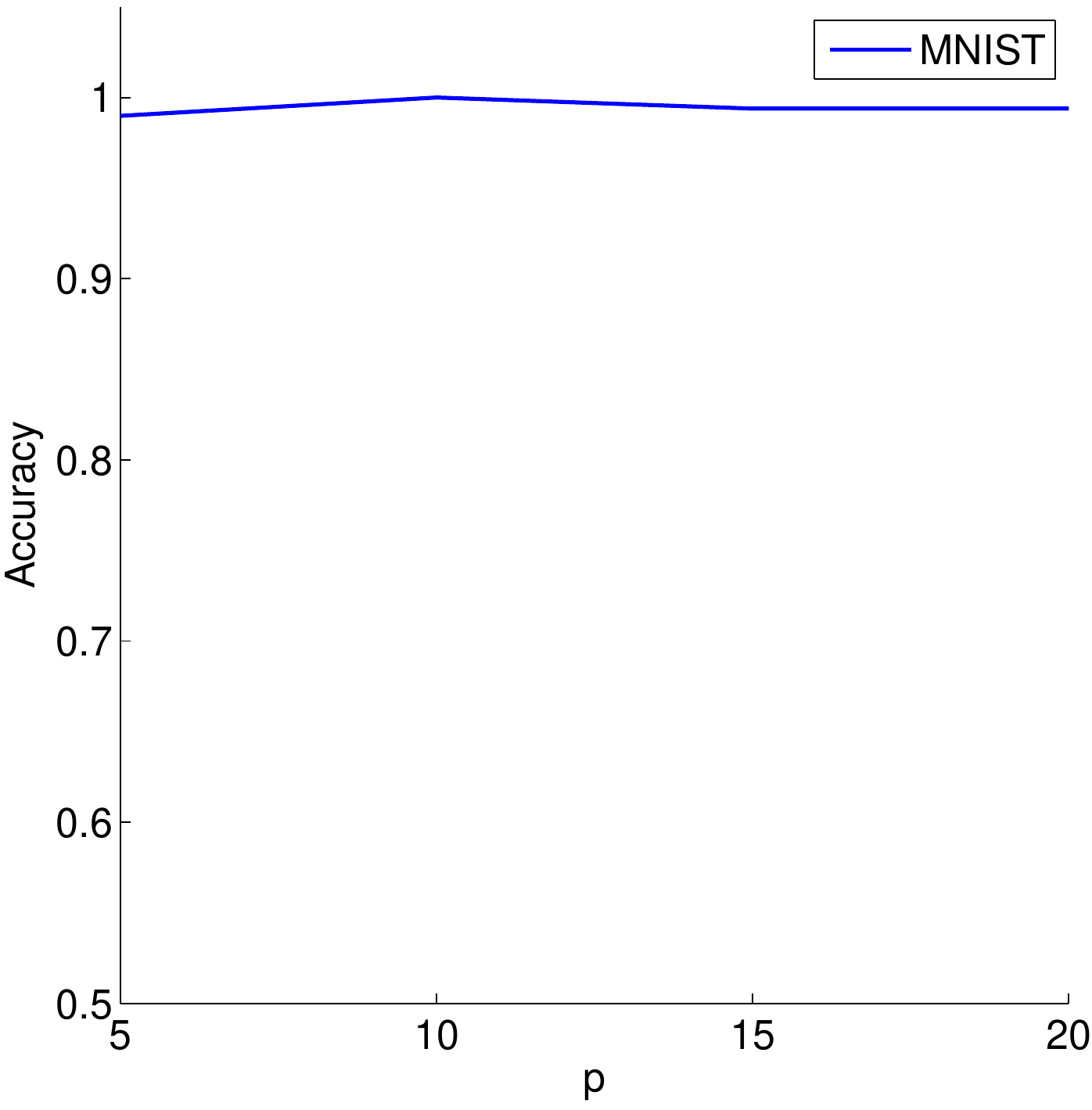}}
\hspace{0.1in}
\subfigure[]{ \label{Orderfig:b} 
\includegraphics[width=0.3\textwidth]{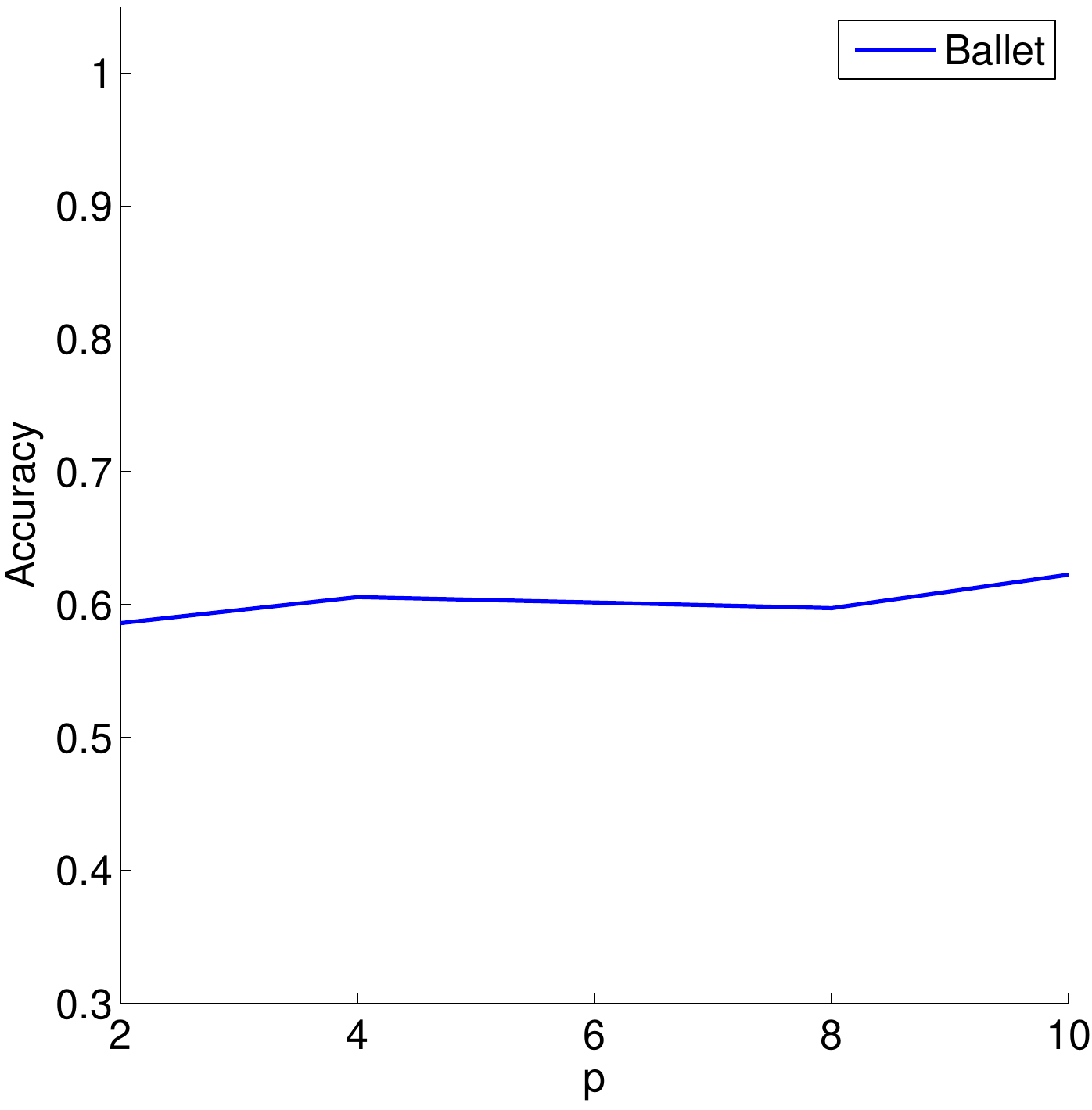}}
\hspace{0.1in}
\subfigure[]{ \label{Orderfig:b} 
\includegraphics[width=0.3\textwidth]{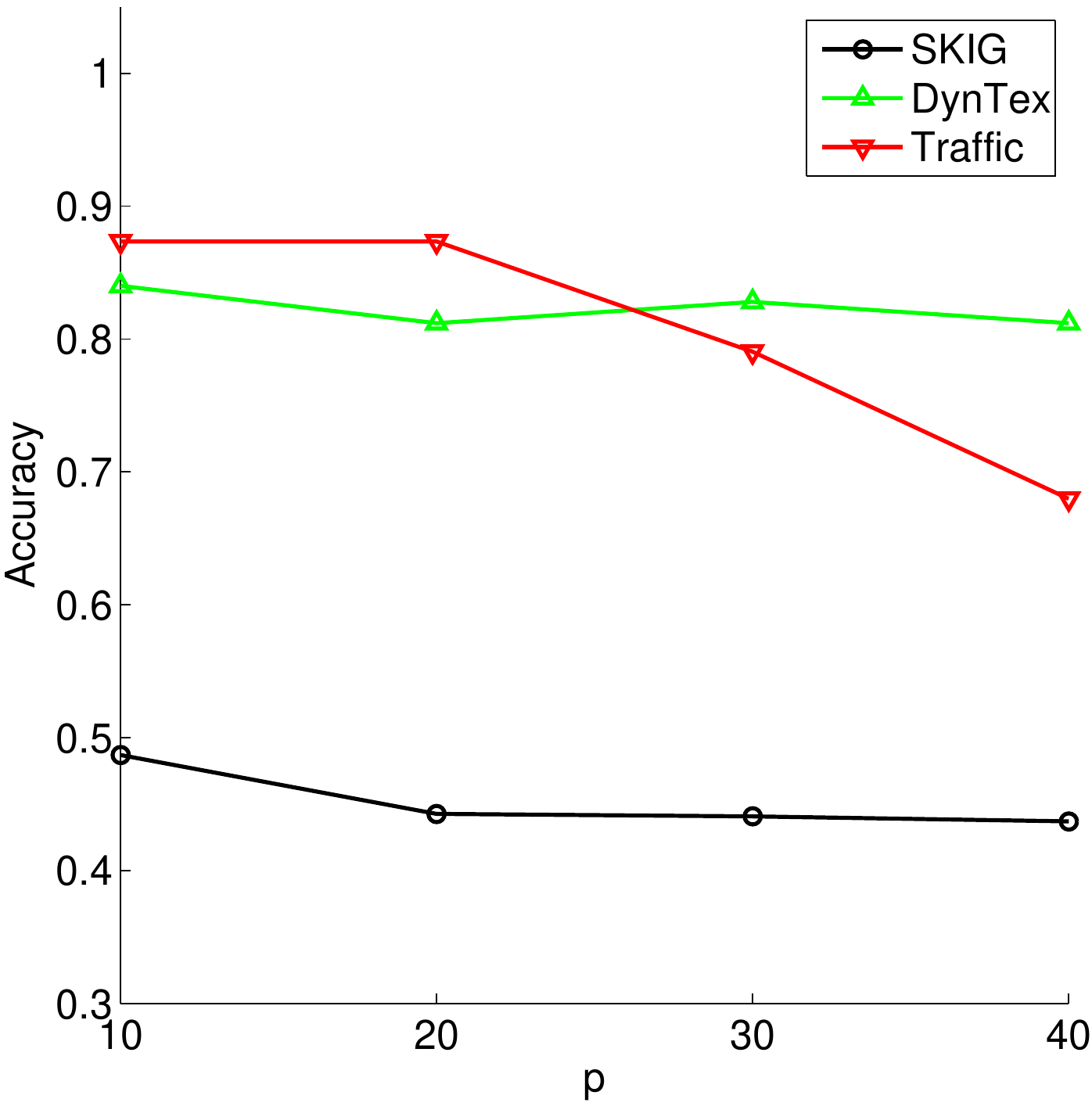}}
\caption{The effects of the order of Grassmann manifold $\mathcal{G}(p,n)$. (a) Clustering accuracy with the various $p$ from $5$ to $20$. (b) Clustering accuracy with the various $p$ from $2$ to $10$ on Ballet dataset. (c) Clustering accuracy with the various $p$ from $10$ to $50$ on rest datasets.} \label{Orderfig}
\end{figure*}

For the order of Grassmann manifold $p$, which is bounded by the number of images in an image set $M$, 
we test its effects on the algorithm performance for all datasets, as shown in Fig. \ref{Orderfig}. These experimental results demonstrate that the order of Grassmann manifold has little impact on the clustering accuracy in most cases, especially for small p. For convenience and fairness, we set $p = 10$ in our experiments except for Ballet dataset where $p = 6$.

The error tolerance $\varepsilon$ is also an important parameter in controlling the terminal condition, which bounds the allowed reconstructed error. We experimentally seek a proper value of $\varepsilon$ to make the iteration process stop at an appropriate level of reconstructed error. Here we set $\varepsilon = 1.0\times 10^{-5}$ for all experiments.

To assess the impact of these parameters, we will conduct experiments by varying one parameter while keeping others fixed to find the best performance parameter values.

\subsubsection{Data Representation for the Compared Methods}
As mentioned in Section \ref{Sec:2}, given an image set $\{\mathbf Y_i\}_{i=1}^M$ where $\mathbf Y_i$ is a grey image with dimension $m\times n$, the corresponding Grassmann point $\mathbf X \in \mathcal{G}(p,d)$ can be constructed. Since the LGLRR, GLRR-F, SLRR, and SCGSM methods are based on the Grassmann manifold, the Grassmann point $\mathbf X\in \mathcal{G}(p,d)$ is naturally taken as the input data for these three methods.

SMCE and LS3C belong to the category of manifold learning methods, which demand a vector form of inputs. The subspace form of points on the Grassmann manifold cannot be used directly, and vectoring the Grassmann point $\mathbf X\in \mathcal{G}(p,d)$ will destroy the geometry of data. For a fair comparison with our method, we map the Grassmann point $\mathbf X\in \mathcal{G}(p,d)$ to a symmetric matrix $\mathbf X \mathbf X^T \in R^{d\times d}$ and then vectorize the symmetric matrix \cite{HarandiSandersonShenLovell2013} as $\text{vec}(\mathbf X\mathbf X^T)$, which becomes the input for the SMCE and LS3C methods.

SSC and LRR are treated as benchmarks since they are classical spectral clustering methods, and, in particular, the LRR method is used in our proposed method. They also need the vector form of data as input. When using $\mathbf X\mathbf X^T$,  this is actually the GLRR-F method \cite{WangHuGaoSunYin2014}. In our experiments, we vectorize the whole image set into a very long vector by stacking all vectors of the raw data in a particular order. However, in most of the experiments, we cannot simply use these long vectors because of high dimensionality for larger image sets. In this case, we apply PCA to reduce these vectors to a low dimension, which is equal to the number of PCA components retaining 95\% of its variance energy. Then PCA-projected vectors will be taken as inputs for SSC and LRR. Finally, we summarize these data presentations for methods compared in Table \ref{Datarepresentationtab}.

\begin{table}
      \centering
   \begin{tabular}{|c|c|}
     \hline
              Methods & Input data for different methods\\
              \hline
              LGLRR &   $\mathbf X\in \mathcal{G}(p,d)$\\
              \hline
              GLRR-F \cite{WangHuGaoSunYin2014} &  $\mathbf X\in \mathcal{G}(p,d)$\\
              \hline
              LRR \cite{LiuLinSunYuMa2013} &  $\text{PCA}(\mathbf Y)$\\
              \hline
              SSC \cite{ElhamifarVidal2013} &  $\text{PCA}(\mathbf Y)$\\
              \hline
              SLRR \cite{YinGaoGuo2015}   &  $\mathbf X\in \mathcal{G}(p,d)$\\
              \hline
              SCGSM \cite{TuragaVeeraraghavanSrivastavaChellappa2011} &  $\mathbf X\in \mathcal{G}(p,d)$\\
              \hline
              SMCE \cite{ElhamifarVidalNips2011} &  $\text{vec}(\mathbf X\mathbf X^T)$\\
              \hline
              LS3C  \cite{PatelNguyenVidal2013} &  $\text{vec}(\mathbf X\mathbf X^T)$\\
     \hline
   \end{tabular}
  \caption{Type of input data for compared methods, where $\mathbf X\in\mathcal{G}$ lies on the Grassmann manifold and $\mathbf X\mathbf X^T$ is the corresponding mapping on the Symmetric metrix space.}\label{Datarepresentationtab}
\end{table}


In our experiments, the performance of different algorithms is measured by the following  clustering accuracy
\[
\text{Accuracy} = \frac{\text{number of correctly classified points}}{\text{total number of points}}\times 100\%.
\]

All the algorithms are coded in Matlab 2014a and implemented on an Intel Core i7-4770K 3.5GHz CPU machine with 32G RAM.

\subsection{Clustering on MNIST Handwirtten Dataset}
\begin{figure}
    \begin{center}
    \includegraphics[width=0.45\textwidth]{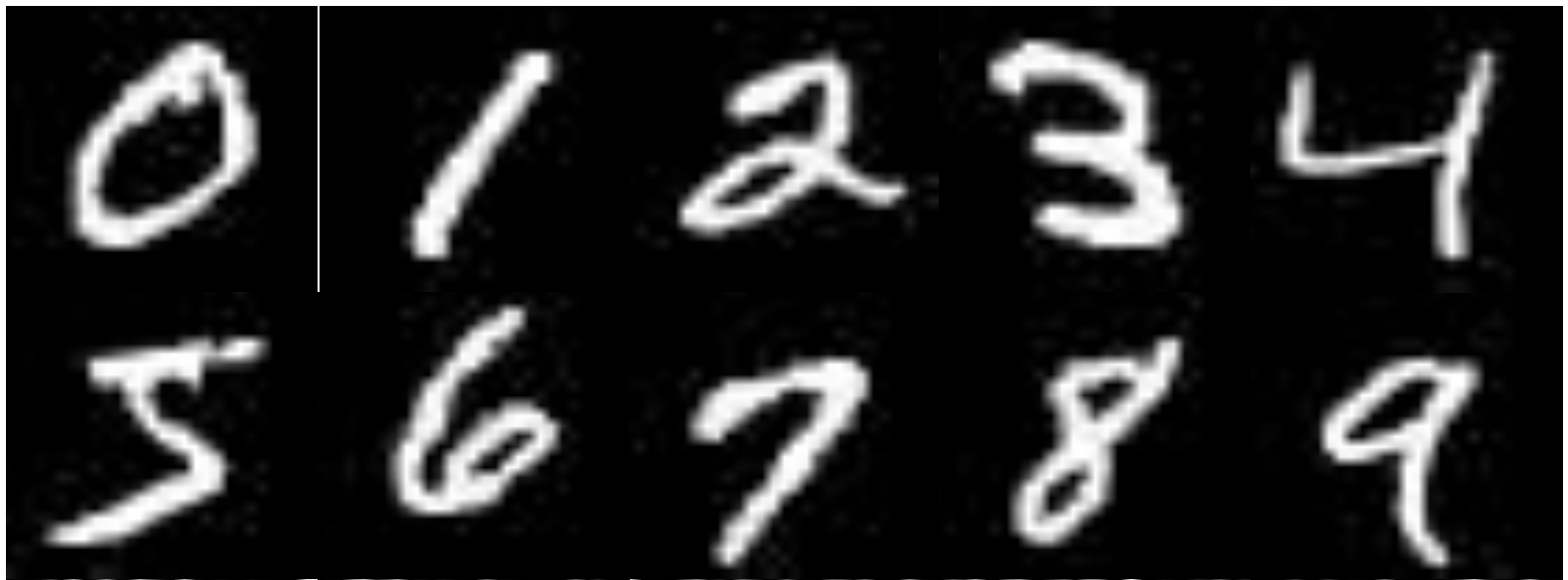}
    \end{center}
    \caption{Some samples from the MNIST handwritten dataset.}\label{FigE1}
\end{figure}

The MNIST dataset has been widely used in testing the performance of pattern recognition algorithms. The digit images in this dataset were written by about 250 volunteers. In the recognition application, around 60,000 digit images in the dataset are used as training data and 10,000 images are used as testing data \cite{LinYangHsiaoChen2015}. All the digit images in this dataset have been size-normalized and centered in a fixed size of $28\times 28$. Fig. \ref{FigE1} shows some digit samples of this dataset.

For each digit, we randomly select $M=20$ images to generate an image set and obtain a total of $N=495$ image sets to test. Note that each image set itself contains only the same digit. The cluster number $R=10$ 
and the dimension of subspace is set $p=10$.  As a result, we construct a Grassmann point $[\mathbf X]\in \mathcal{G}(10,784)$ for each image set.
For the SSC and LRR methods, we vectorize each image set and reduce the dimension of each image set $28\times28\times20=15680$ to 383 by PCA. In this experiment, we set the parameters $\lambda = 1$ and $C=50$.

As there exists a clear difference among the different digit samples in this dataset, the performance should be relatively better. 
In fact, Table \ref{MNISTtab} shows that almost all methods achieve great performance. 

\begin{table}
   \centering
   \begin{tabular}{|c|c|c|c|}
     \hline
              \diagbox{Methods}{Dataset} & MNIST \\
              \hline
              LGLRR &  1\\
              \hline
              GLRR-F \cite{WangHuGaoSunYin2014}&  1\\
              \hline
              LRR \cite{LiuLinSunYuMa2013}   &  0.9010\\
              \hline
              SSC \cite{ElhamifarVidal2013}   &  0.8242\\
              \hline
              SLRR \cite{YinGaoGuo2015}   &  0.9737\\
              \hline
              SCGSM \cite{TuragaVeeraraghavanSrivastavaChellappa2011}&  0.8727\\
              \hline
              SMCE \cite{ElhamifarVidalNips2011}&  1\\
              \hline
              LS3C  \cite{PatelNguyenVidal2013} &  0.9152\\
     \hline
   \end{tabular}
  \caption{Experimental results on the MNIST handwritten dataset.}\label{MNISTtab}
\end{table}

\subsection{Clustering on Action Datasets}
The Grassmann manifold is a good tool for representing image sets, so it is appropriate to use it to represent video sequence data, which is regarded as an image set. In this experiment, we select two challenge action video datasets, the ballet dataset and SKIG action clips, to test the proposed method's performance. The ballet dataset, which contains simple backgrounds, could verify the capacity of the proposed method for action recognition in an ideal condition; while SKIG, which has more variations in background and illumination, examines the robustness of the proposed method for noise. We also explore the effect of the neighborhood size $C$ and the balance parameter $\lambda$ in these challenge experiments.

\subsubsection{Ballet Dataset}
This dataset contains 44 video clips, collected from an instructional ballet DVD. The dataset consists of eight complex action patterns performed by three subjects. These eight actions include: ``left-to-right hand opening'', ``right-to-left hand opening'', ``standing hand opening'', ``leg swinging'', ``jumping'', ``turning'', ``hopping'', and ``standing still''. The dataset is challenging due to the significant intra-class variations in terms of speed, spatial and temporal scale, clothing, and movement. The frame images are normalized and centered in a fixed size of $30 \times 30$. Some frame samples of ballet dataset are shown in Fig.~\ref{FigE3}.

Similar to the method that constructs image sets for this dataset in \cite{ShiraziHarandiSandersonAlaviLovell2012}, we split each clip into sections of $M=12$ frames and each section is taken as an image set. We obtain in total $N=713$ image sets. The cluster number is $R=8$ and the dimension of subspace is set to $p=6$. Thus we construct Grassmann points $[\mathbf X]\in \mathcal{G}(6,900)$ for clustering. For the SSC and LRR methods, the subspace vectors at $30\times 30\times 12 = 10800$  are reduced to the  dimension of $135$ by PCA. We set $\lambda=15$ and $C=90$.

Table \ref{Ballettab} presents the experimental results of all the algorithms on the ballet dataset. 
As the ballet images do not have a very complex background or other obvious disturbs, they can be regarded as clean data without noise. Additionally, the images in each image set have time sequential relations and each action consists of several simple actions. So these help to improve the performances of the evaluated methods, as shown in Table \ref{Ballettab}. Our method, GLRR-F, SLRR, SCGSM, SMCE, and LS3C are obviously superior to other methods, and this reflects the advantages of using manifold information.

\begin{figure*}
\centering
\subfigure[]{ \label{Balletfig:a} 
\includegraphics[width=0.30\textwidth]{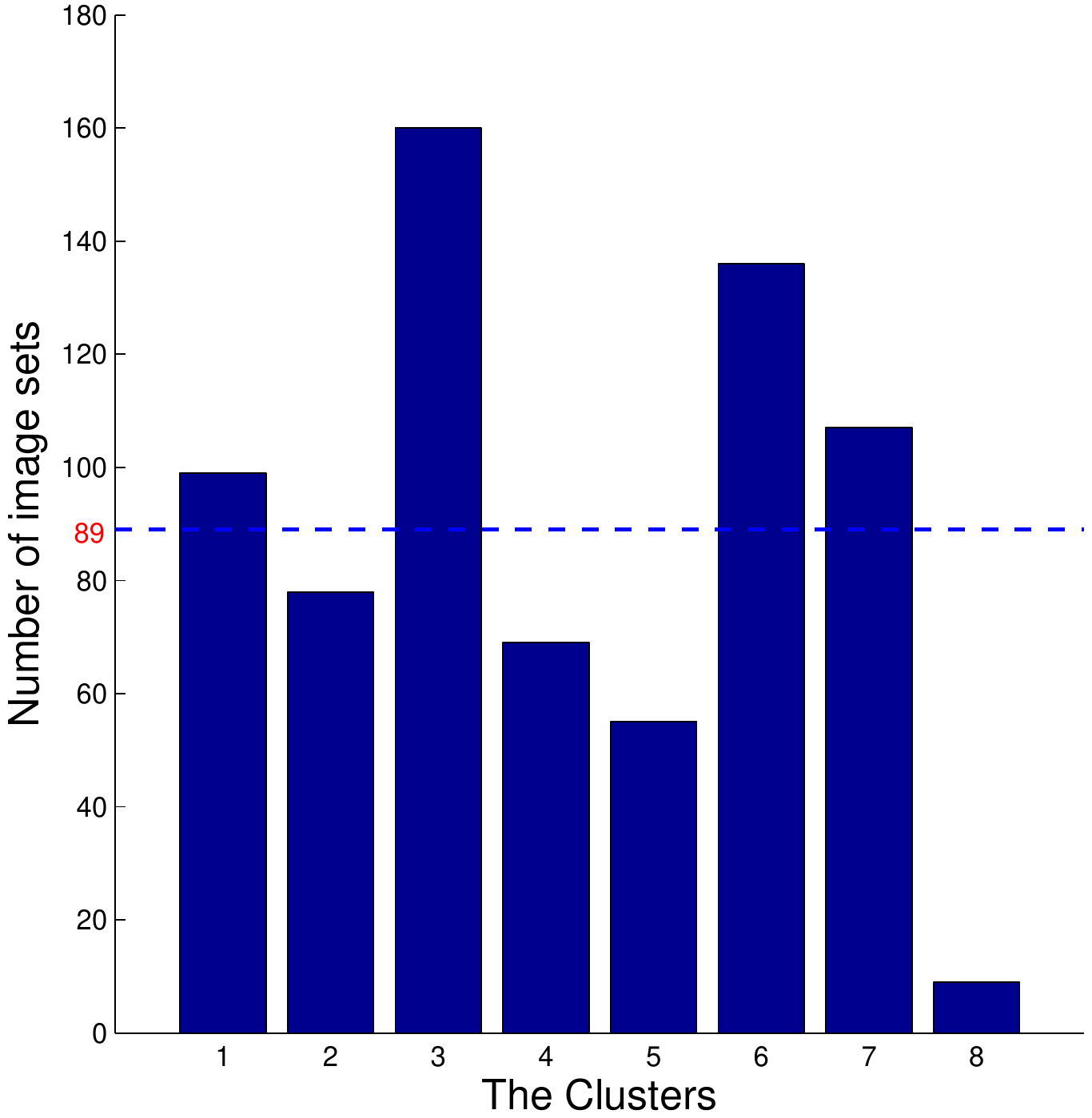}}
\hspace{0.1in}
\subfigure[]{ \label{Balletfig:a} 
\includegraphics[width=0.30\textwidth]{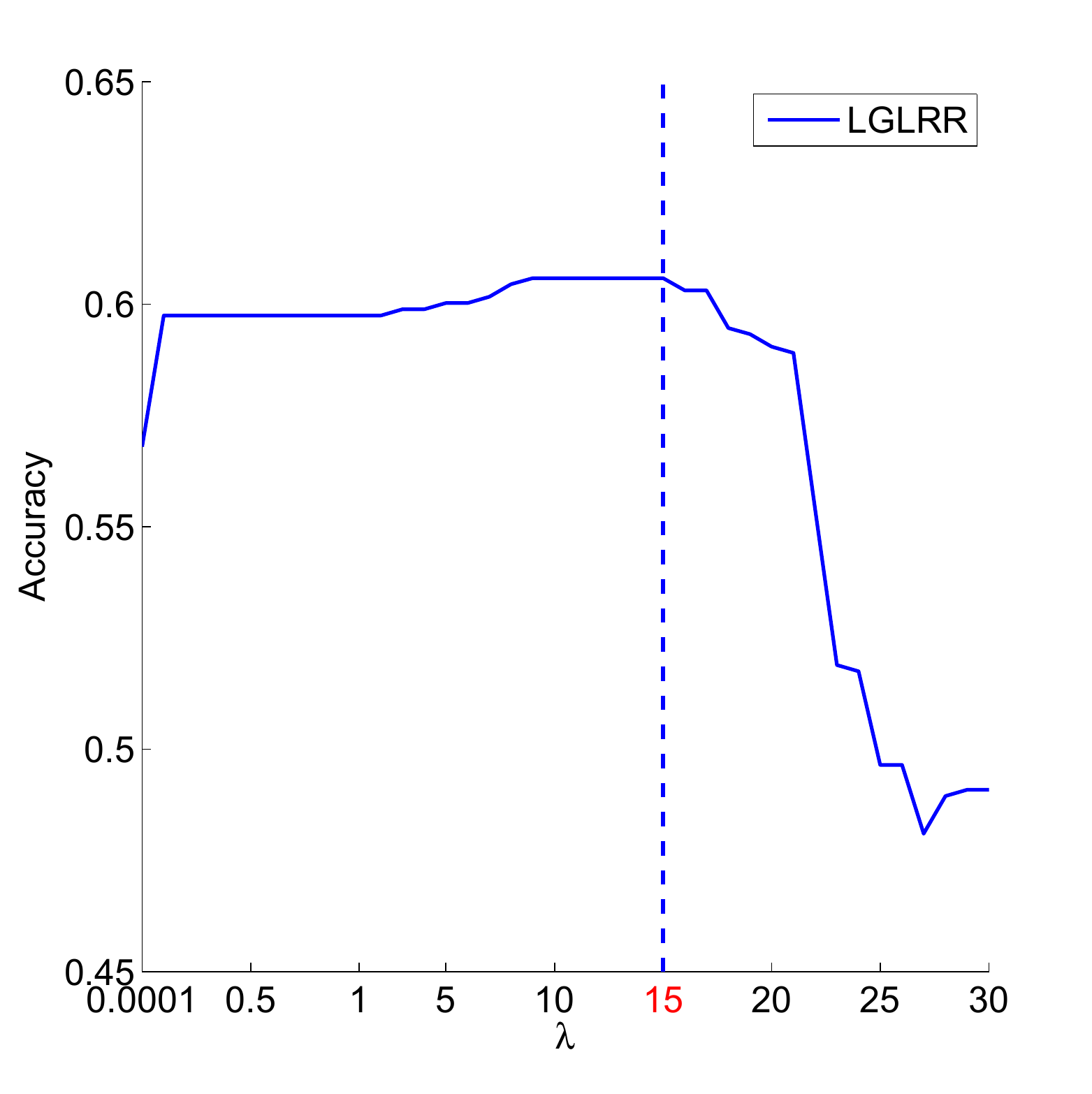}}
\hspace{0.1in}
\subfigure[]{ \label{Balletfig:b} 
\includegraphics[width=0.30\textwidth]{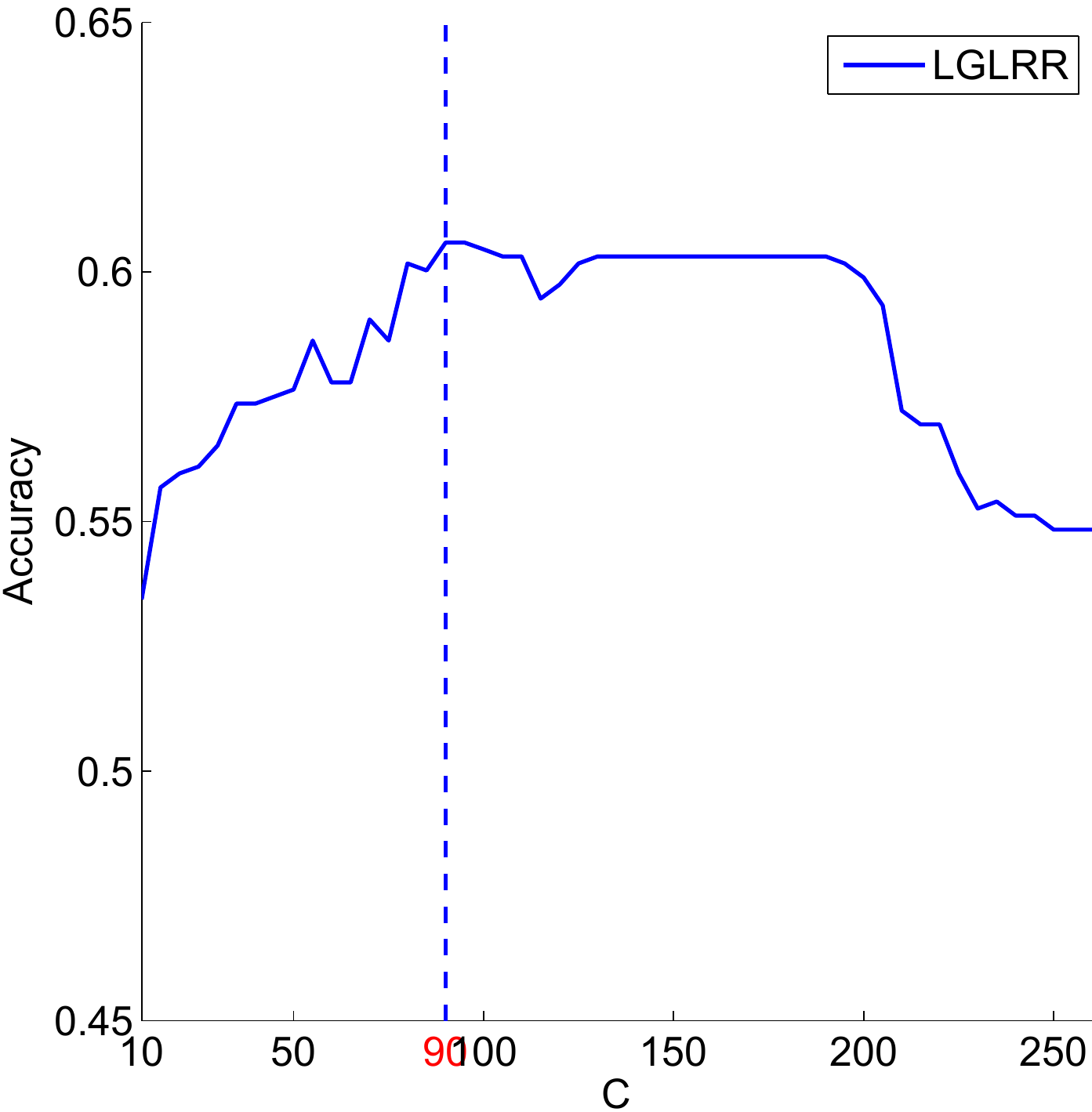}}
\caption{Parameter selections in ballet experiment. (a) Number of image sets in each cluster of the test data; (b) Clustering accuracy with the parameter $\lambda$ varying; (c) Clustering accuracy with different neighborhood sizes $C$.} \label{Balletfig}
\end{figure*}

Fig. \ref{Balletfig}(a) shows the number of image sets in each cluster. Now we examine the parameter sensitivity of the proposed method. In the proposed method, $\lambda$ is used to balance the corruption and the rank of representation coefficients, and $C$ is used to control the sparsity of representation coefficients. We vary alternatively the parameters $\lambda$ and $C$ to test the performance of the proposed method. Fig. \ref{Balletfig}(b) and Fig. \ref{Balletfig}(c) respectively show the changes of accuracy versus parameters $\lambda$ and $C$. We fix $C=90$ in \ref{Balletfig}(b), and we can see that the proposed method maintains the highest accuracy in a wide range of $\lambda$ values from 9 to 15. We fix $\lambda=15$ in Fig. \ref{Balletfig}(c). The proposed method achieves the highest performance when the neighborhood size $C$ is between $85$ to $90$, which is actually the average number of image sets in different clusters as shown in Fig. \ref{Balletfig}(a).

\subsubsection{SKIG Dataset}
\begin{figure}
    \begin{center}
    \includegraphics[width=0.45\textwidth]{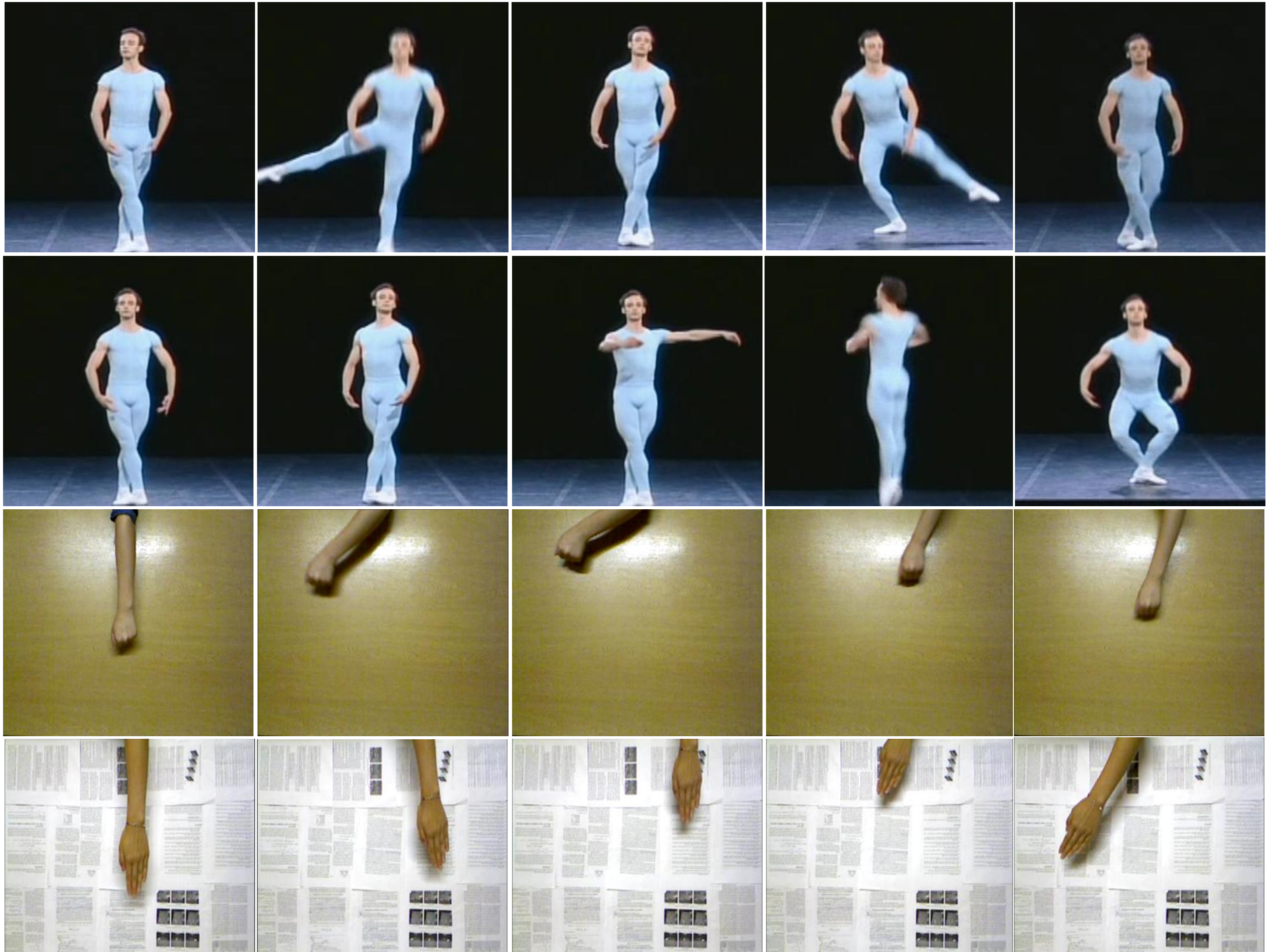}
    \end{center}
    \caption{Some samples from the ballet dataset and the SKIG dataset. The top two rows are from the ballet dataset and the bottom two rows are from the SKIG dataset.}\label{FigE3}
\end{figure}

This dataset contains 1080 RGB-D sequences captured by a kinect sensor. This dataset stores ten kinds of gesture of six persons: ``circle'', ``triangle'', ``up-down'', ``right-left'', ``wave'', ``Z'', ``cross'', ``comehere'', ``turn-around'', and ``pat''. All the gestures are performed by fist, finger, and elbow respectively with three backgrounds (wooden board, white plain paper, and paper with characters) and two illuminations (strong light and poor light). Each RGB-D sequence contains 63 to 605 frames. Here the images are normalized to $24\times 32$ with mean zero and unit variance. Fig. \ref{FigE3} shows some samples of RGB images.

We regard each RGB video as an image set and obtain in total $N=540$ image sets for clustering into $R=10$ classes. The dimension of subspace is set $p=10$ and thus the Grassmann points on $\mathcal{G}(10,768)$
are constructed. Since there is a big gap between 63 and 605 frames among SKIG sequences, and the PCA algorithm requires each sample have an equal dimension, it is difficult to select the same number of frames for each sequence as the inputs for SSC and LRR. Thus, we cease comparing our method with SSC and LRR. We set $\lambda=7$ and $C=65$.

However, this dataset is more challenging than the ballet dataset, due to the smaller scale of the objects, the various backgrounds, and illuminations. From the experimental results in Table \ref{Ballettab}, we can conclude that the proposed method is superior to other methods.

\begin{figure*}
\centering
\subfigure[]{ \label{SKIGfig:a} 
\includegraphics[width=0.32\textwidth]{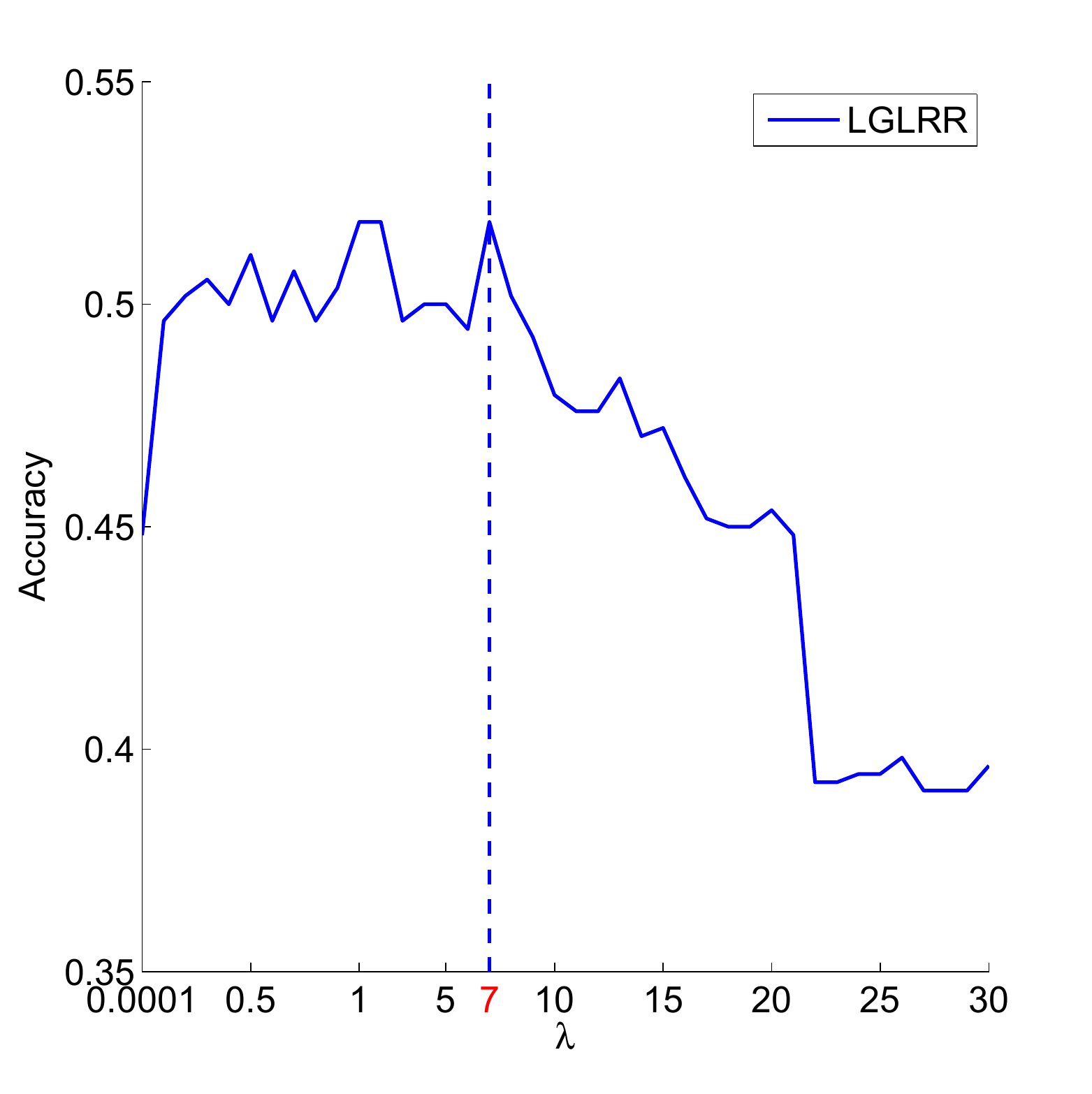}}
\hspace{0.1in}
\subfigure[]{ \label{SKIGfig:b} 
\includegraphics[width=0.32\textwidth]{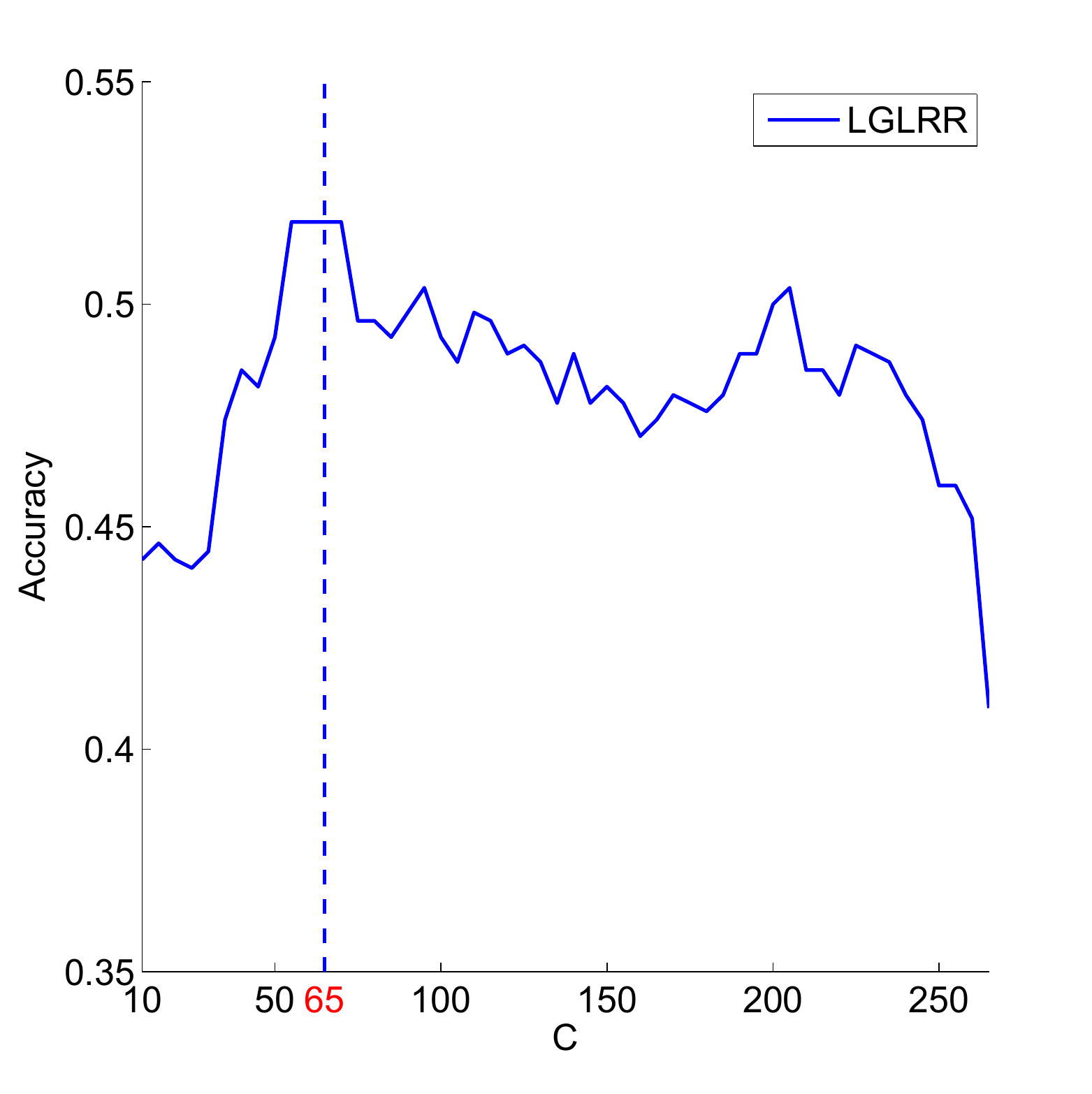}}
\caption{Parameter selection in the SKIG experiment. (a) Clustering accuracy with the various $\lambda$ values; (b) Clustering accuracy with different neighborhood sizes $C$.} \label{SKIGfig} 
\end{figure*}

We also implement experiments with different parameters to get the proper parameters setting. Fig. \ref{SKIGfig}(a) shows that the accuracy obtained from the proposed method decreases when $\lambda$ is larger than $7$. Compared with the results for the Ballet dataset, we select a relative small $\lambda=6.5$. As shown in Fig. \ref{SKIGfig}(b), the highest accuracy is obtained between $C = 55$ and $C = 70$. Therefore, the parameter $C$ is set as the average number of image sets of clusters; here it is fixed at 65 for each cluster. 
These experimental results verify the effects of low rank and localized constraints in our proposed methods again.

\begin{table}
   \centering
   \begin{tabular}{|c|c|c|}
     \hline
              \diagbox{Methods}{Datasets} & Ballet & SKIG\\
              \hline
              LGLRR & 0.6087 & 0.5185\\
              \hline
              GLRR-F \cite{WangHuGaoSunYin2014}& 0.5905 & 0.5056\\
              \hline
              LRR \cite{LiuLinSunYuMa2013} & 0.2819 & -\\
              \hline
              SSC \cite{ElhamifarVidal2013}& 0.2903 & -\\
              \hline
              SLRR \cite{YinGaoGuo2015} & 0.5316 & 0.3907\\
              \hline
              SCGSM \cite{TuragaVeeraraghavanSrivastavaChellappa2011}& 0.5877 & 0.3704\\
              \hline
              SMCE \cite{ElhamifarVidalNips2011}& 0.5105  & 0.4611\\
              \hline
              LS3C \cite{PatelNguyenVidal2013}& 0.4222  & 0.4148\\
     \hline
   \end{tabular}
  \caption{Subspace clustering results on the ballet and SKIG dataseta.}\label{Ballettab}
\end{table}

\subsection{Clustering on the DynTex++ Dataset}
\begin{figure}[!h]
    \begin{center}
    \includegraphics[width=0.45\textwidth]{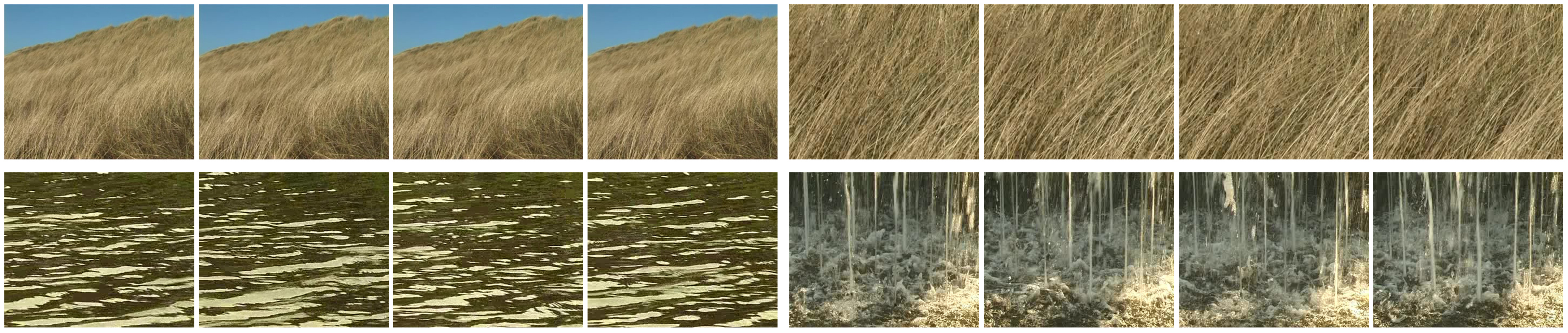}
    \end{center}
    \caption{Some samples from the DynTex++ dataset.}\label{FigE2a}
\end{figure}

The DynTex++ dataset is derived from a total of 345 video sequences in different scenarios, including river water, fish swimming, smoke, clouds, and so on. These videos are labeled according to 36 classes and each class has 100 subsequences (3,600 subsequences in total) with a fixed size of $50\times 50\times 50$. Some DynTex++ samples are shown in Fig. \ref{FigE2a}.

In this experiment, we want to verify the performances of the proposed method on various cluster numbers $R$ from 3 to 10. For the experiments, we randomly select 50 videos for each cluster. We treat each video as an image set and pick up the dimension of subspace $p=10$. For the SSC and LRR methods, we reduce the dimension of a video $50\times50\times50=125,000$ to $\{27, 43, 78, 68, 77, 87, 98, 97\}$ by PCA with respect to various cluster numbers from $3$ to $10$. Here we set $\lambda=1$ and the neighborhood size as $C=50$.

\begin{figure}
    \begin{center}
    \includegraphics[width=0.45\textwidth]{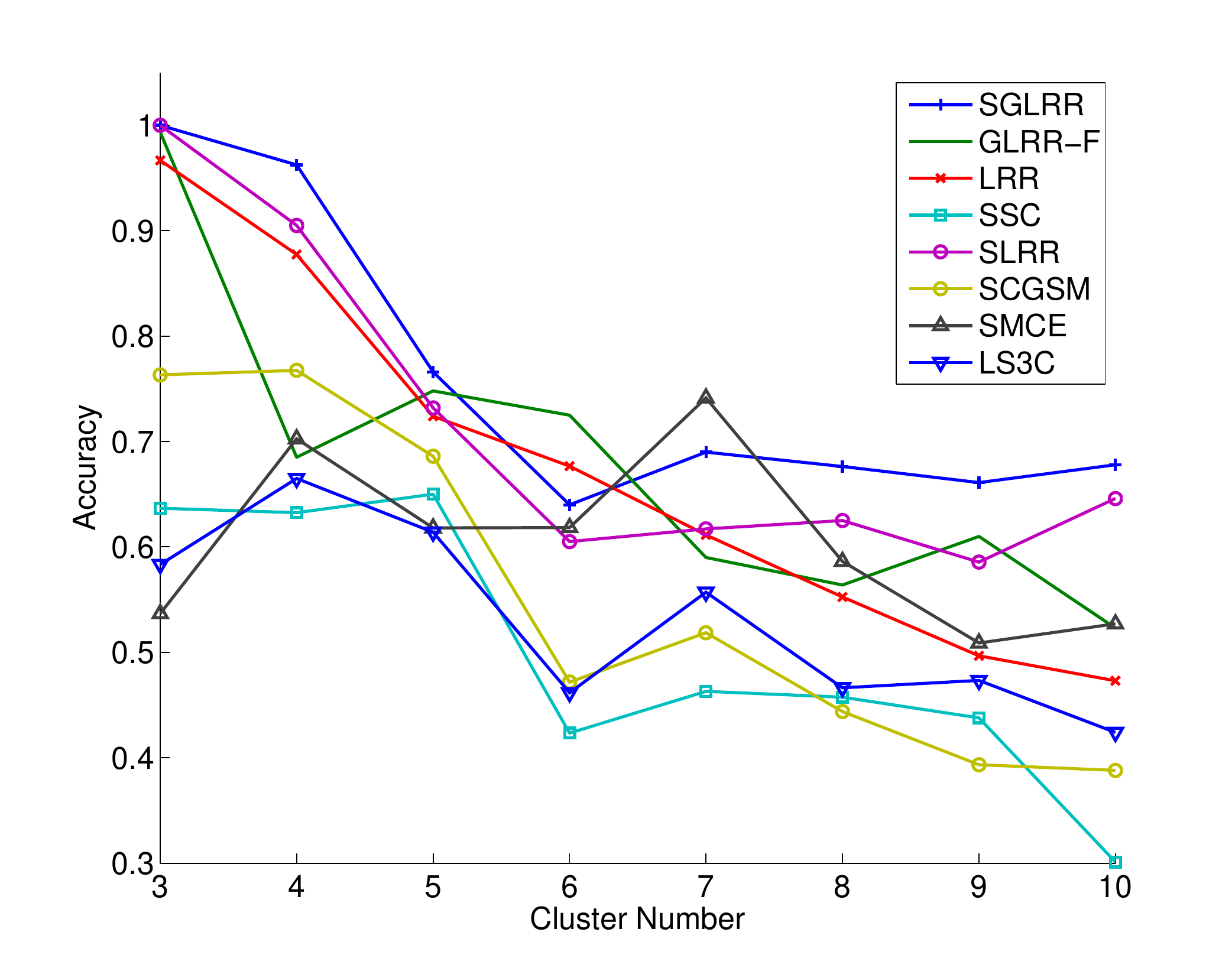}
    \end{center}
    \caption{Experimental result of the DynTex++ dataset.}\label{FigE2b}
\end{figure}

This is a challenging dataset for clustering because most of the textures from different classes is fairly similar. Fig. \ref{FigE2b} presents the clustering results of all methods while varying the number of classes. It is obvious that the  accuracy of the proposed method is superior to the other methods in most cases. We also observe that our proposed method is quite robust to changing the number of classes, compared with other methods. However, the accuracy of the proposed method decreases when the cluster number $R$
is between 4 and 6.  This may be caused by the clustering challenge when more similar texture images are added to the datasets.

\subsection{Clustering on the Traffic Dataset}
\begin{figure}
    \begin{center}
    \includegraphics[width=0.45\textwidth]{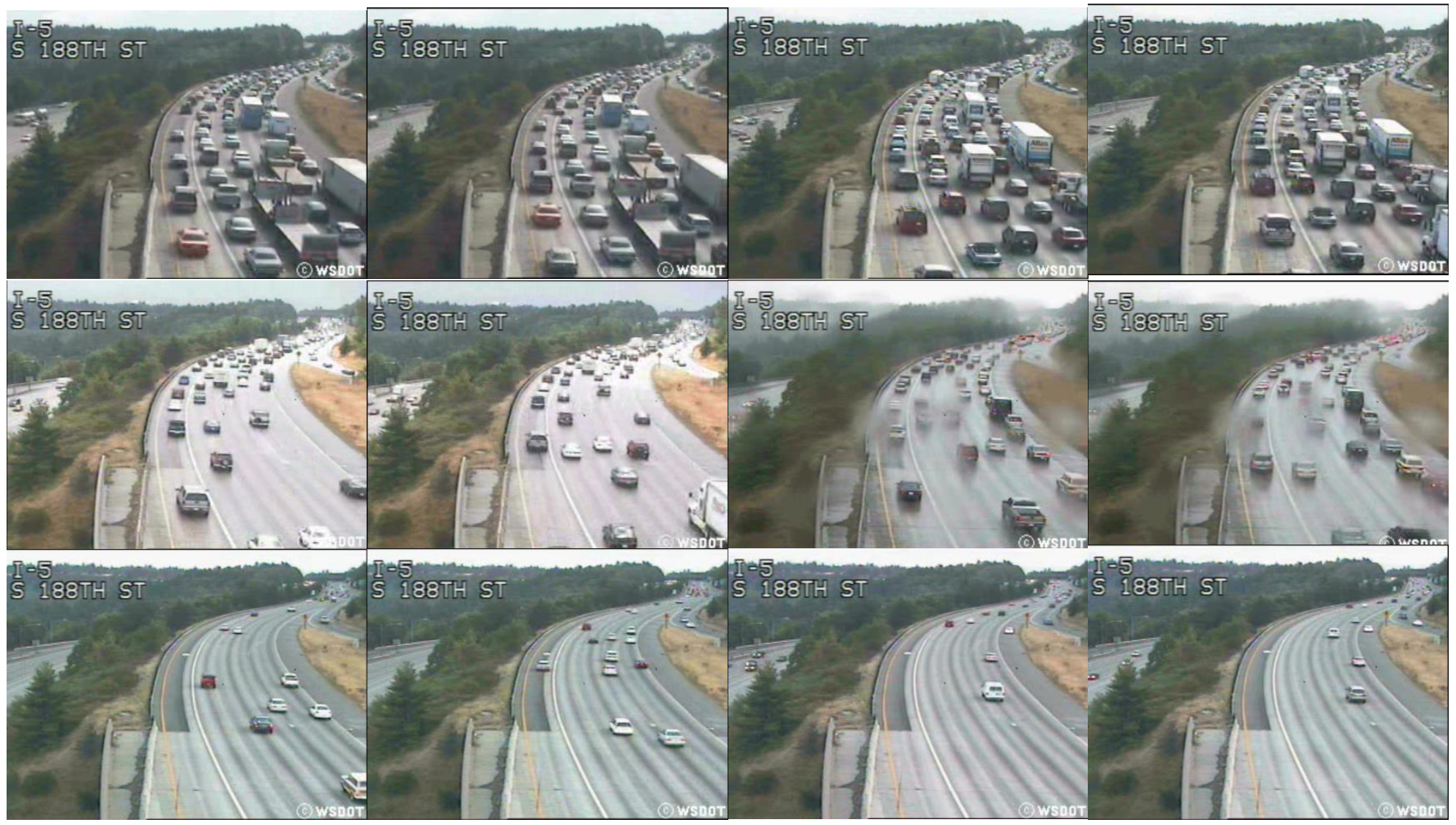}
    \end{center}
    \caption{Some samples from the highway traffic dataset.}\label{FigE5}
\end{figure}

In this experiment, we inspect the proposed method on practical applications with more complex conditions, such as the traffic dataset. The traffic dataset used in this experiment contains 253 video sequences of highway traffic captured under various weather conditions, such as sunny, cloudy, and rainy. These sequences have three traffic levels: light, medium and heavy. There are 44 heavy clips, 45 medium clips, and 164 light clips. Each video sequence has 42 to 52 frames. The video sequences are converted to gray images and each image is normalized to $24 \times 24$ with mean zero and unit variance. Some samples of the highway traffic dataset are shown in Fig.~\ref{FigE5}.

We regard each video sequence as an image set to construct a point on the Grassmann manifold as in the previous experiments. The subspace dimension is set as $p=10$ and the number of clusters equals the number of traffic levels, i.e., $R=3$. For SSC and LRR, we vectorize the first 42 frames in each clip (discarding the rest of the frames in the clip) and then use PCA to reduce the dimension $24\times24\times42=24192$ to 147. Here we set $\lambda = 1.5$ and the neighbor size as $C=61$. Note that the traffic jam level doesn't have a sharp borderline. For some clips, it is difficult to say whether they belong to the heavy, medium, or light level. So clustering is a challenging task.

The experimental results are listed in Table \ref{Traffictab}. Obviously, the proposed method and GLRR-F outperform other methods by at least 19\%, which almost reaches the accuracy of some classification algorithms \cite{SankaranarayananTuragaBaraniukChellappa2010}. Like the Dyntex++ dataset, frames in each video of the traffic dataset changes slowly, so the performance of SSC and LRR is not bad. As the frames in the ballet dataset consist of rapid action changes in each video, the performance of SSC and LRR is poor. This phenomenon justifies that the proposed method is robust to the changes inside the image sets.
\begin{table}
   \centering
   \begin{tabular}{|c|c|}
     \hline
              \diagbox{Methods}{Datasets} & Traffic\\
              \hline
              LGLRR &  0.8735\\
              \hline
              GLRR-F \cite{WangHuGaoSunYin2014}& 0.8498 \\
              \hline
              LRR \cite{LiuLinSunYuMa2013} & 0.6838 \\
              \hline
              SSC \cite{ElhamifarVidal2013}& 0.6285 \\
              \hline
              SLRR \cite{YinGaoGuo2015} & 0.6087 \\
              \hline
              SCGSM \cite{TuragaVeeraraghavanSrivastavaChellappa2011}& 0.6443 \\
              \hline
              SMCE \cite{ElhamifarVidalNips2011}& 0.5613 \\
              \hline
              LS3C \cite{PatelNguyenVidal2013}& 0.6364 \\
     \hline
   \end{tabular}
  \caption{Subspace clustering results on the traffic dataset.}\label{Traffictab}
\end{table}
\section{Conclusion}\label{Sec:6}
In this paper, building on Grassmann manifold geometry and Log mapping, the classic LRR method is extended for subspace objects on Grassmann manifold. The data self-expression used in LRR is realized over the local tangent spaces at points on the Grassmann manifold, thus a novel localized LRR on Grassmann manifold, namely LGLRR, is presented. An efficient algorithm is proposed to solve  the LGLRR model. The algorithm is tested on a number of image sets and video clip datasets, and the experiments confirm that LGLRR is very suitable for representing non-linear high-dimensional data and revealing intrinsic multiple subspace structures in clustering applications. The experiments also demonstrate that the proposed method is robust in a variety of complicated practical scenarios. As part of future work, we will explore the LRR method on the Grassmann manifold in the intrinsic view.

\section*{Acknowledgements}
The research project is supported by the Australian Research Council (ARC) through grant DP140102270 and also partially supported by the National Natural Science Foundation of China under Grant No. 61390510, 61133003, 61370119 and  61227004, Beijing Natural Science Foundation No. 4132013 and 4162010, Project of Beijing Educational Committee grant No. KM201510005025 and Funding PHR-IHLB of Beijing Municipality.


\bibliographystyle{IEEEtran}
\bibliography{reference_boyue}


\begin{IEEEbiography}[{\includegraphics[width=1in,height=1.25in,clip,keepaspectratio]{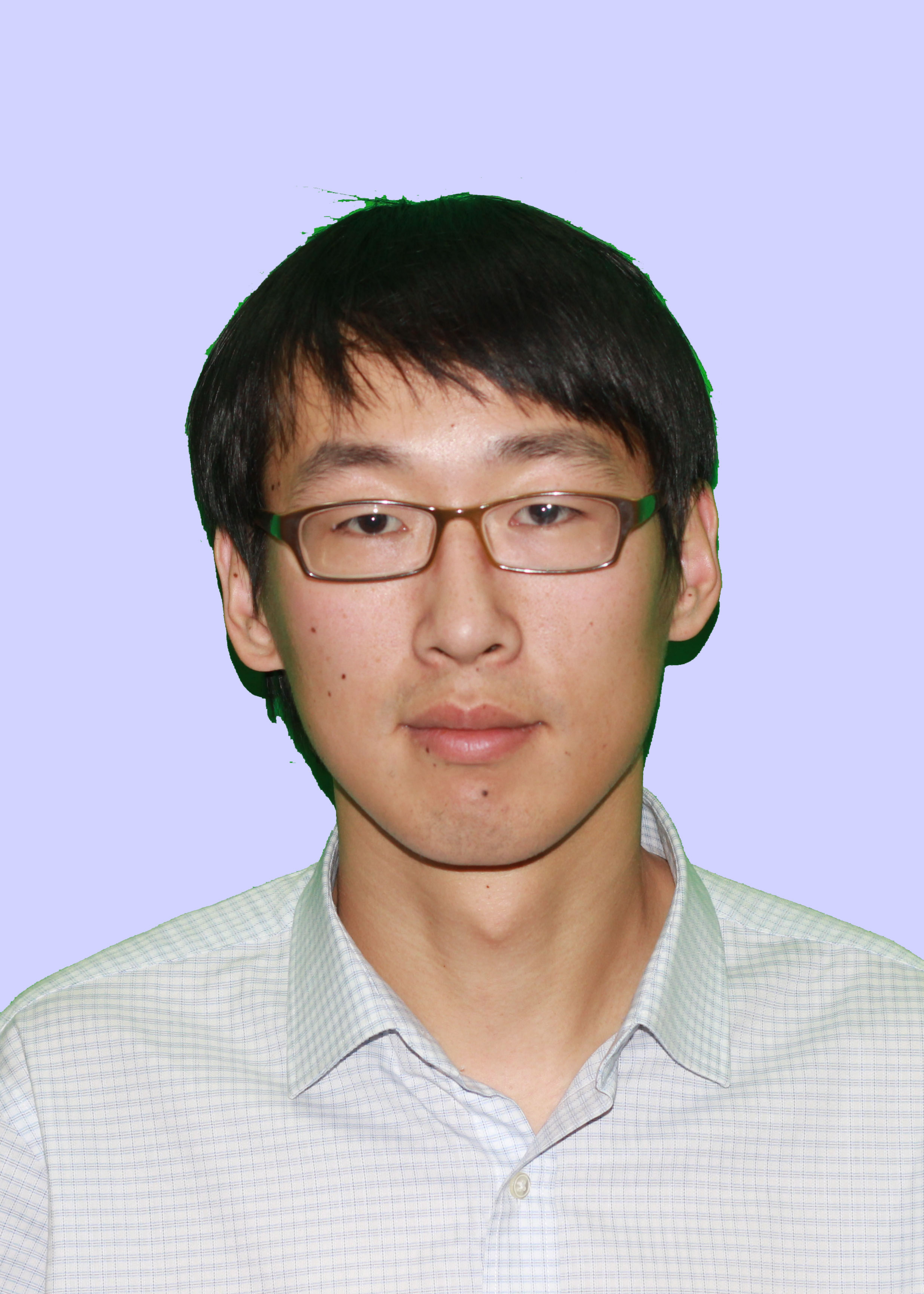}}]
{Boyue Wang} received his B.Sc. from Hebei University of Technology,
Tianjin, China, in 2012. He is currently pursuing Ph.D. at the Beijing Municipal Key Laboratory of Multimedia and Intelligent Software Technology,
Beijing University of Technology, Beijing.
His current research interests include computer
vision, pattern recognition, manifold learning, and kernel methods.
\end{IEEEbiography}

\begin{IEEEbiography}[{\includegraphics[width=1in,height=1.25in,clip,keepaspectratio]{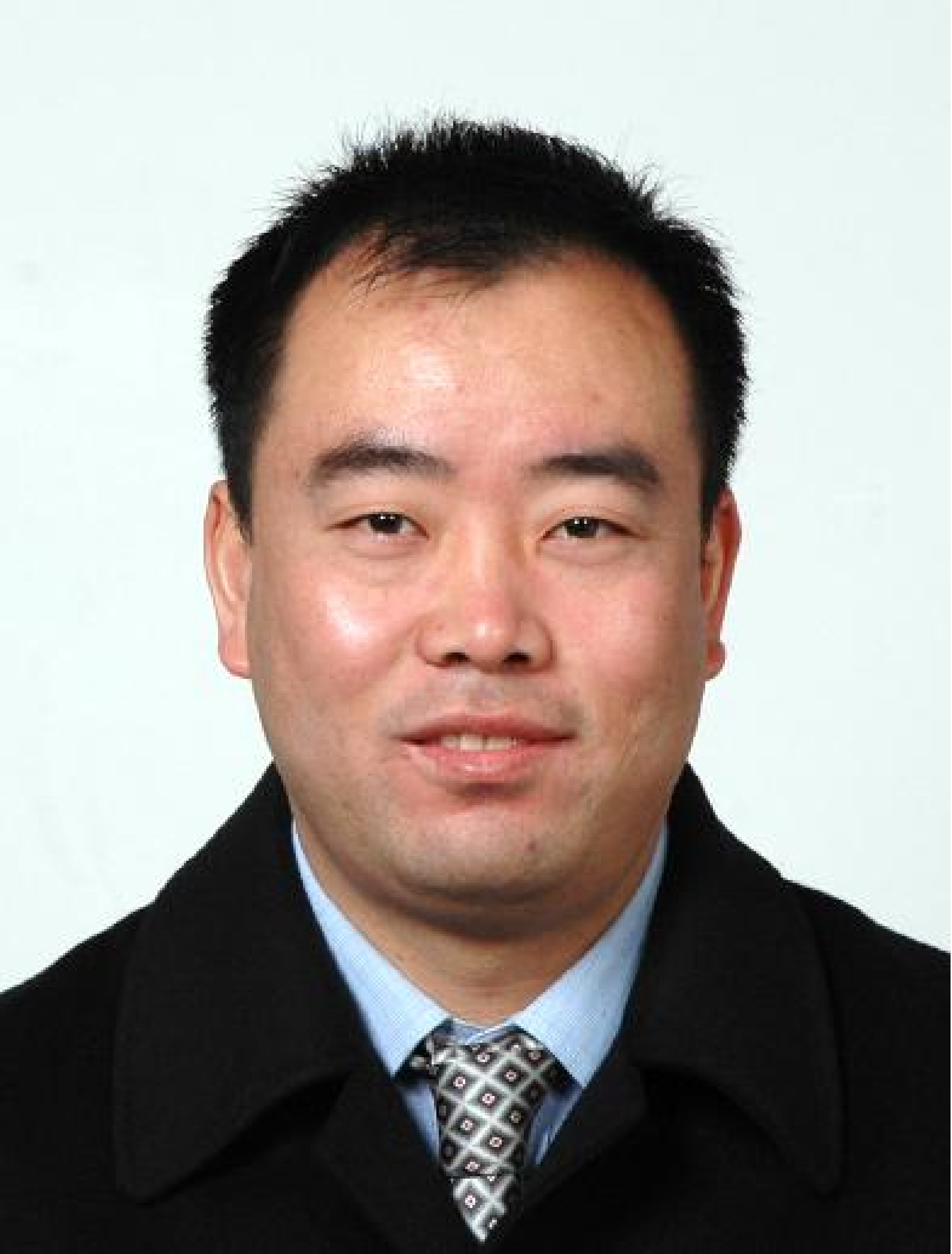}}]
{Yongli Hu} received his Ph.D. from Beijing University of Technology in 2005. He is a professor at College of Metropolitan Transportation at Beijing University of Technology. He is
a researcher at the Beijing Municipal Key Laboratory of Multimedia and Intelligent Software Technology.
His research interests include computer graphics, pattern recognition, and multimedia technology.
\end{IEEEbiography}

\begin{IEEEbiography}[{\includegraphics[width=1in,height=1.25in,clip,keepaspectratio]{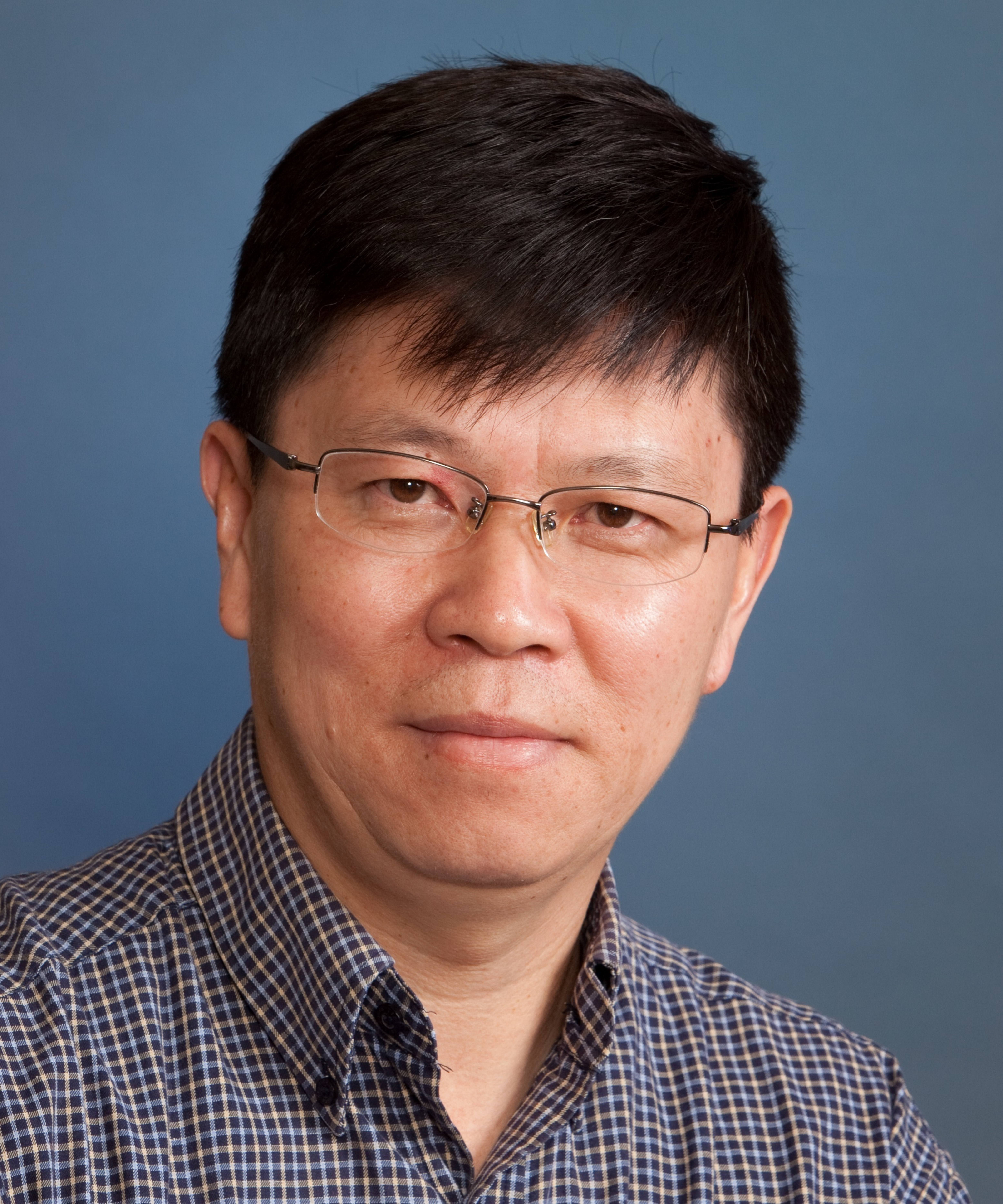}}]
{Junbin Gao} graduated from Huazhong University of Science and Technology (HUST),
China in 1982 with a BSc. in Computational Mathematics and
obtained his PhD. from Dalian University of Technology, China in 1991. He is a Professor of Big Data Analytics in the University of Sydney Business School at the University of Sydney and was a Professor in Computer Science
in the School of Computing and Mathematics at Charles Sturt
University, Australia. He was a senior lecturer, a lecturer in Computer Science from 2001 to 2005 at the
University of New England, Australia. From 1982 to 2001 he was an
associate lecturer, lecturer, associate professor, and professor in
Department of Mathematics at HUST. His main research interests
include machine learning, data analytics, Bayesian learning and
inference, and image analysis.
\end{IEEEbiography}

\begin{IEEEbiography}[{\includegraphics[width=1in,height=1.25in,clip,keepaspectratio]{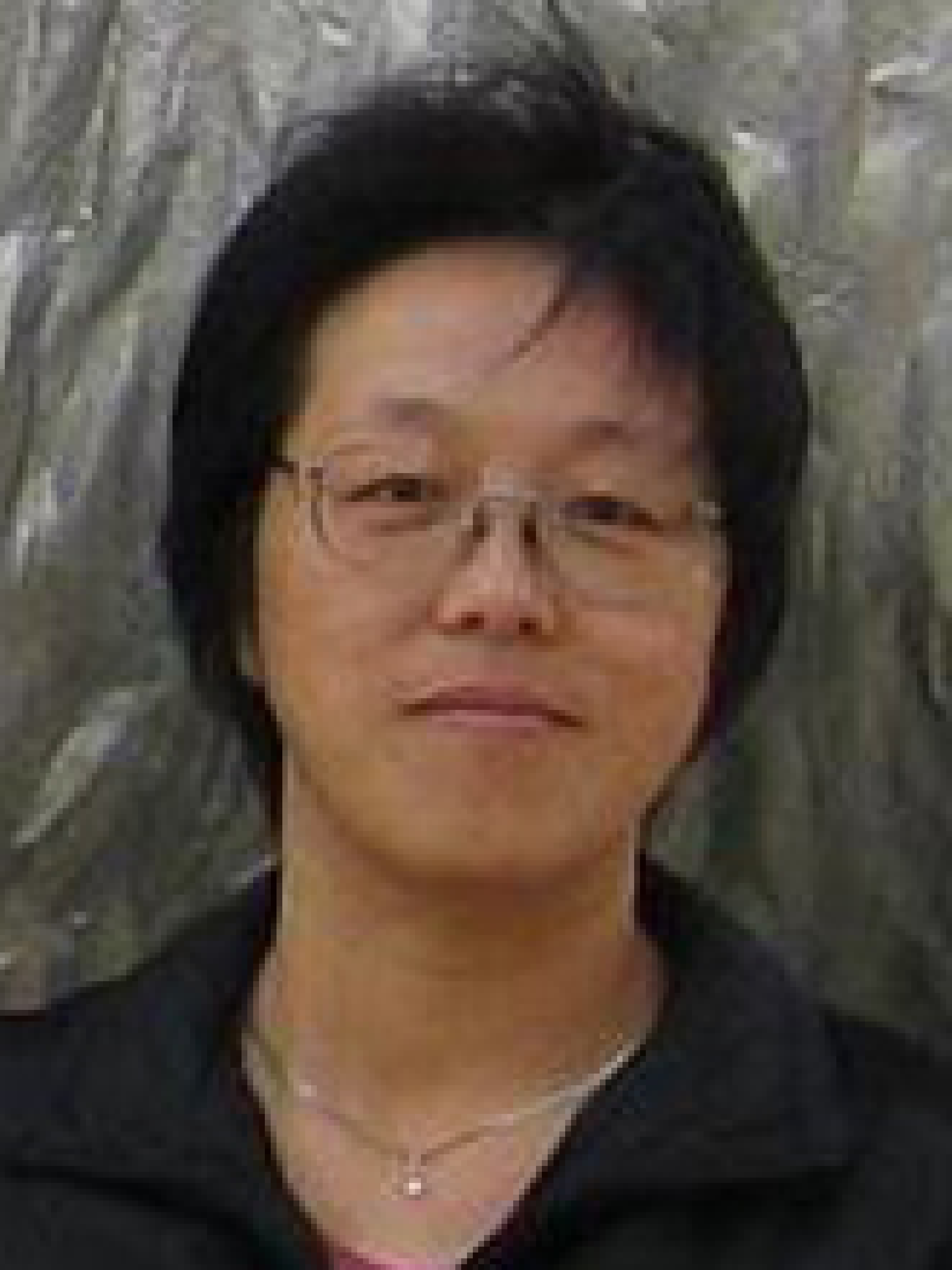}}]
{Yanfeng Sun} received her Ph.D. from Dalian University of Technology in 1993. She is a professor at College of Metropolitan Transportation at Beijing University of Technology. She is
a researcher at the Beijing Municipal Key Laboratory of Multimedia and Intelligent Software Technology. She is a member of the China Computer Federation.
 Her research interests are multi-functional perception and image processing.
\end{IEEEbiography}

\begin{IEEEbiography}[{\includegraphics[width=1in,height=1.25in,clip,keepaspectratio]{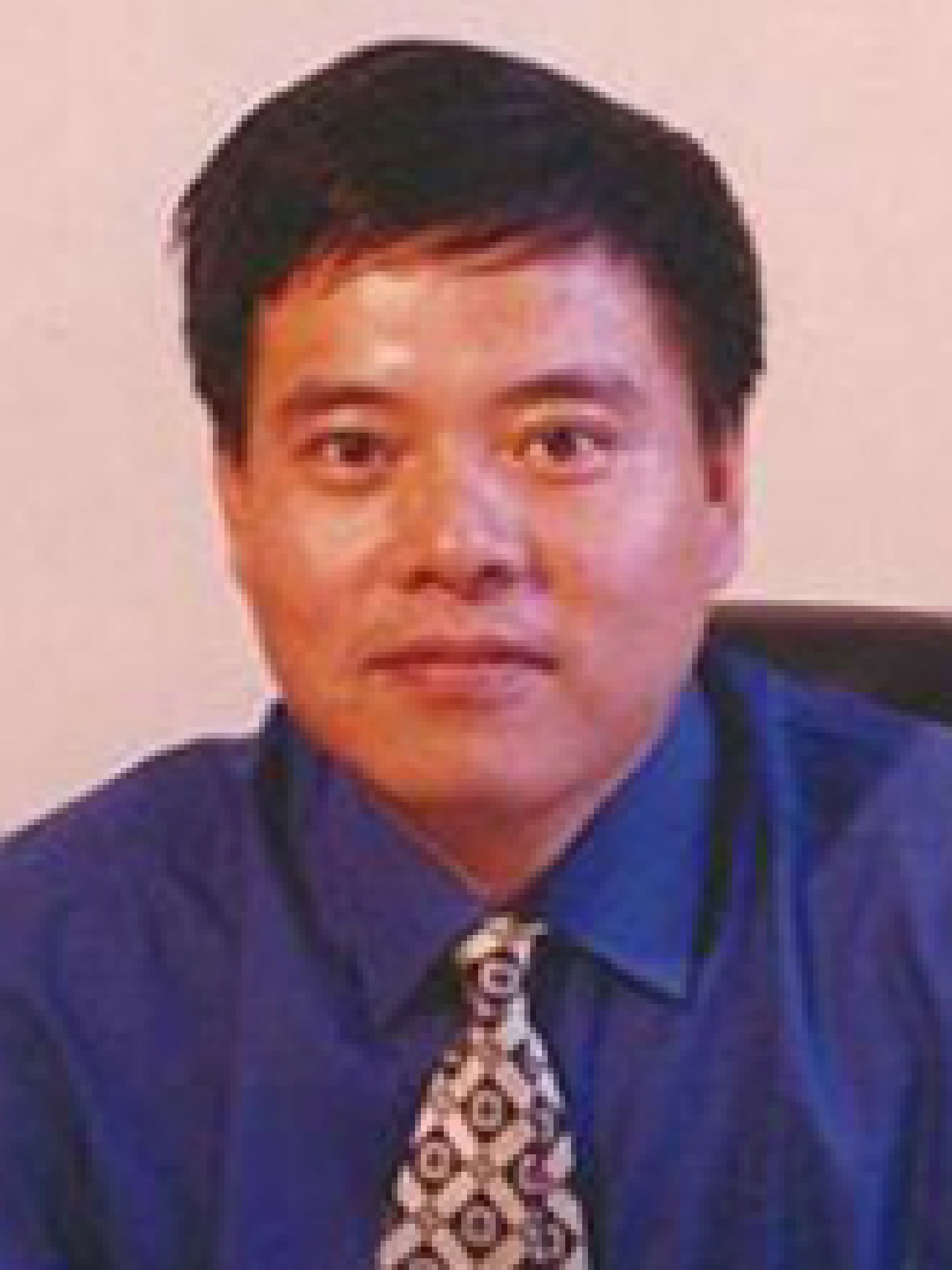}}]
{Baocai Yin} received his Ph.D. from Dalian University of Technology in 1993.
He is a Professor in the College of Computer Science and Technology, Faculty of Electronic Information and Electrical Engineering, Dalian University of Technology. He is
a researcher at the Beijing Municipal Key Laboratory of Multimedia and Intelligent Software Technology.
He is a member of the China Computer Federation. His
research interests include multimedia, multifunctional perception, virtual reality, and computer graphics.
\end{IEEEbiography}
\vfill

\end{document}